\documentclass[a4paper,fleqn]{cas-sc}

\usepackage[authoryear]{natbib}
\usepackage{longtable}
\usepackage{tabularx}
\usepackage{adjustbox}
\usepackage{float}
\usepackage{algorithm}
\usepackage{algpseudocode}
\usepackage{algorithmicx}
\usepackage{longtable}
\usepackage{pdflscape}
\usepackage{array}
\usepackage{booktabs}
\usepackage{multirow}
\usepackage{changepage}

\def\tsc#1{\csdef{#1}{\textsc{\lowercase{#1}}\xspace}}
\tsc{WGM}
\tsc{QE}
\begin{document}
\let\WriteBookmarks\relax
\def\floatpagepagefraction{1}
\def\textpagefraction{.001}

\shorttitle{Digital Twins of Natural Ecosystems}    

\shortauthors{Coro and Bove}  

\title [mode = title]{Building Conversational Digital Twins of Natural Ecosystems with Reactive Artificial Intelligence}




\author[1]{Gianpaolo Coro}[orcid=0000-0001-7232-191X]

\cormark[1]


\ead{gianpaolo.coro@cnr.it}

%\ead[url]{https://www.isti.cnr.it/it/chi-siamo/people-detail/78/Gianpaolo_Coro}

\credit{Conceptualization, Methodology, Software, Validation, Formal analysis, Resources, Supervision, Funding acquisition, Writing - Original Draft, Writing - Review \& Editing}

\affiliation[1]{organization={Istituto di Scienza e Tecnologie dell'Informazione ``A. Faedo'' del Consiglio Nazionale delle Ricerche (CNR-ISTI)},
            addressline={via Moruzzi, 1}, 
            city={Pisa},
%          citysep={}, % Uncomment if no comma needed between city and postcode
            postcode={56124}, 
            %state={Italy},
            country={Italy}}

\author[1]{Pasquale Bove}[orcid=0000-0002-9969-8255]

% Footnote of the second author

% Email id of the second author
%\ead{pasquale.bove@cnr.it}

%\ead[url]{https://www.isti.cnr.it/it/chi-siamo/people-detail/873/Pasquale_Bove}

\credit{Conceptualization, Methodology, Software, Validation, Formal analysis, Writing - Review \& Editing}


\cortext[1]{Corresponding author}



\begin{abstract}
%A framework for building a conversational digital twin of a natural ecosystem (DTE) is presented, implementing a \textit{reactive} architecture that separates scientific computation from language generation, in contrast to autonomous \textit{agentic} systems. The framework integrates harmonised and live data, scientific literature, species distributions, ecosystem risk assessment, and scenario simulation. A case study of the Massaciuccoli Lake basin (Tuscany, Italy) compares the responses of two agentic and four reactive DTEs based on the following large language models (LLMs): Llama-3.2-3B and Gemini-3.5-Flash in both architectures, and Llama-4-Scout and Qwen3-8B in the reactive architecture. Reactive DTEs demonstrated higher correctness, quantitative grounding, and hallucination control, whereas agentic DTEs produced richer ecological narratives. Reactive Gemini obtained the highest quality evaluation scores. Reactive Llama-3.2 approached the performance of larger models, indicating that external reasoning can partially compensate for limited LLM capacity. The framework supports reactive architectures as a transferable, explainable, and cost-effective foundation for conversational DTEs.
A framework for building a conversational digital twin of a natural ecosystem (DTE) is presented. Its \textit{reactive} architecture uses semantic routing and predefined analytical workflows to separate scientific computation from language generation. The framework integrates harmonised and live data, scientific literature, species distributions, ecosystem risk assessment, and scenario simulation. A Massaciuccoli Lake basin case study (Tuscany, Italy) compares two LLM-led retrieval baselines, termed \textit{agentic} in this study, with four reactive configurations: Llama-3.2-3B and Gemini-3.5-Flash in both architectures, plus reactive Llama-4-Scout and Qwen3-8B. Reactive configurations achieved higher correctness, quantitative grounding, and hallucination control, whereas agentic baselines produced richer ecological narratives. Reactive Gemini scored highest overall. Reactive Llama-3.2 approached larger models' performance, suggesting that external analytical computation can partially compensate for limited LLM capacity. The framework provides a modular, explainable foundation for conversational DTEs.
\end{abstract}

\begin{graphicalabstract}
 \includegraphics[width=1\linewidth]{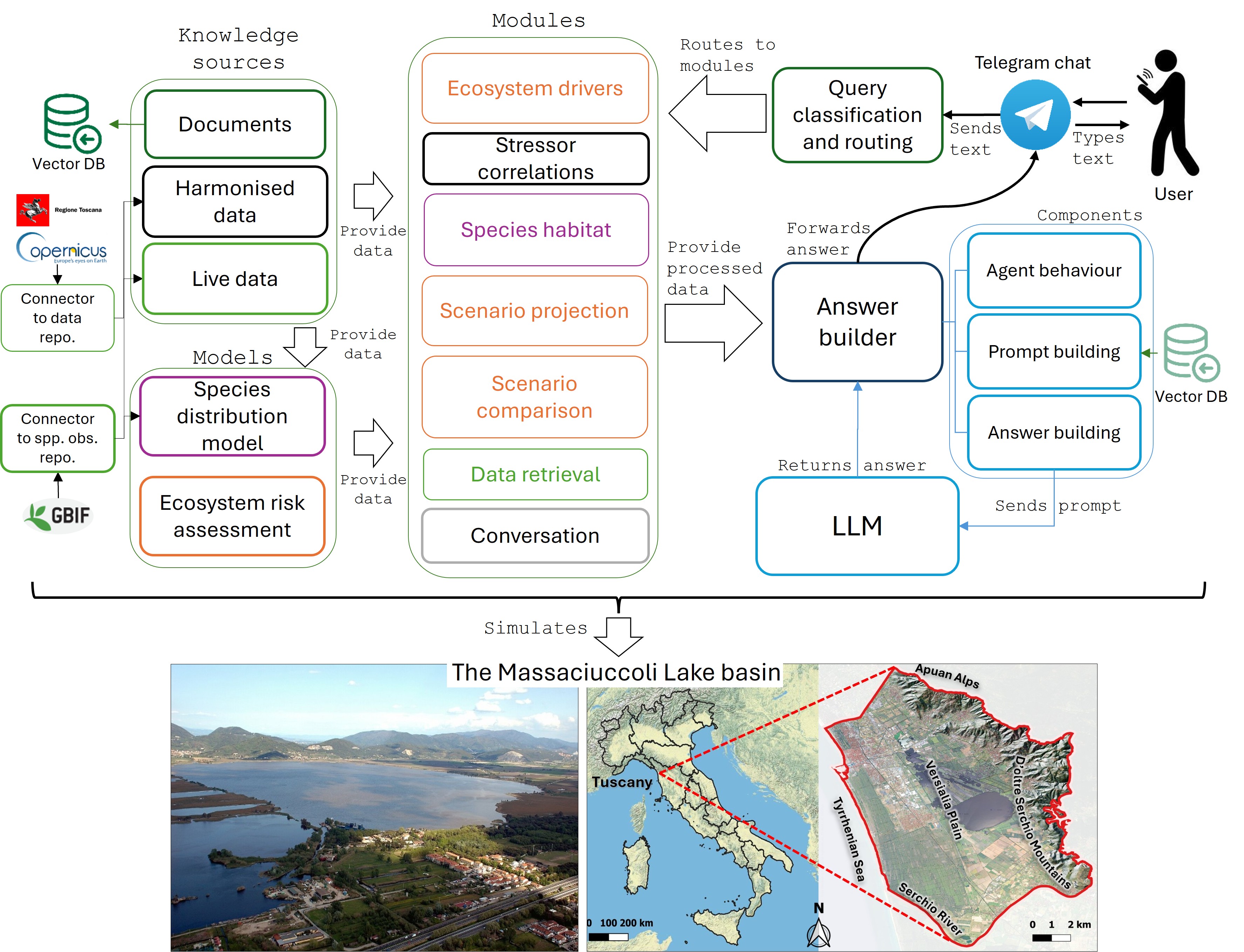}
\end{graphicalabstract}

\begin{highlights}
\item Digital twin of a natural ecosystem based on reactive agents
\item A reactive architecture separates scientific computation from language generation
\item The framework integrates ecosystem data, models, documents, and live observations
\item Reactive agents improve correctness, grounding, and hallucination control
\item Structured reasoning enables small language models to rival larger models

\end{highlights}


\maketitle

% Main text
\section{Introduction}

Ecosystems are increasingly subjected to interacting climatic, environmental, and anthropogenic pressures, with combined effects stemming from complex interactions among variables or processes rather than from individual factors in isolation \citep{bastos2023joint, kefford2023cumulative}. Factors such as climate change, land-use changes, water scarcity, pollution, biological invasions, and resource exploitation collectively impact biodiversity, ecosystem functioning, and the services essential to human communities \citep{mooney2009biodiversity, moustakas2025climate, naderi2025ecosystem}. Interactions often occur across different spatial and temporal scales, leading to nonlinear responses, including critical transitions, that are challenging to predict from fragmented observations. Consequently, environmental management requires computational systems capable of integrating diverse evidence, reconstructing the current state of the ecosystem, exploring potential trajectories, and assessing the effects of various interventions.

Digital twins (DTs) offer a promising framework for addressing this need \citep{blair2021digital,de2023digital,feng2025cognitive}. Among the possible definitions available (Section \ref{sec:overview}), the digital twin of a natural ecosystem (DTE) can be defined as a dynamic virtual counterpart of a real ecosystem that integrates observations, models, simulations, and computational services to represent its state and investigate its evolution over time. Unlike a static geographical representation or the output of an individual model, it maintains a continuous or periodic connection with the physical ecosystem through data streams originating from satellites, \textit{in situ} sensors, monitoring networks, citizen observations, and other information sources. These data are combined with physical, biological, ecological, and data-driven models to estimate variables that cannot be directly observed, identify anomalous conditions, generate projections, and support the comparison of management scenarios. Consequently, a DTE is both a modelling infrastructure and a decision-support environment through which scientists, public authorities, and other stakeholders can access and interpret evidence about ecosystem dynamics. 

A DTE can also adopt a \textit{holistic} perspective \citep{campana2021role}. An ecosystem consists of multiple interdependent subsystems involving its morphology, geology, physical and chemical conditions, biological communities, climatic drivers, human activities, and socioeconomic context. A holistic DTE thus accounts for broad-scale processes such as ecosystem energetics, nutrient cycling, population and community dynamics, biodiversity change, and the production of ecosystem services. Moreover, it should simulate the relationships between these processes and external pressures. In this view, the ecosystem is modelled as a system of systems whose components interact across space and time. The purpose of a holistic DTE is to connect each subsystem sufficiently to investigate how a perturbation can propagate through the ecosystem and produce ecological, economic, or social consequences \citep{coro2020open}.

DTE designs can be generally distinguished into \textit{top-down} and \textit{bottom-up} approaches. In a \textit{top-down} approach, ecosystem components, processes, and interactions are defined in advance based on established ecological theory and expert knowledge, and subsequently translated into explicit mechanistic or conceptual models \citep{rs13112119,de2023digital,lecarpentier2024}. However, a comprehensive \textit{top-down} representation of all ecosystem components and processes is generally impractical. Mechanistic ecosystem models require extensive domain knowledge, assumptions about functional relationships, and substantial effort for parameterisation and calibration. Their development may be slow, and their applicability can be constrained by the spatial and temporal scales for which they were designed \citep{geary2020guide,vedder2024developing}. Moreover, the information required to model every relevant interaction is rarely available with uniform coverage or quality. Environmental observations are collected by different organisations for different purposes, follow heterogeneous standards, and vary substantially in resolution, temporal extent, uncertainty, and accessibility. As the number of components increases, uncertainty propagates through the modelling chain, while the computational and conceptual complexity of integrating the models may become prohibitive. A top-down representation should therefore be regarded as a long-term objective rather than a condition that can be completely achieved in a single implementation \citep{tsakiridis2023cognitive,guidani2024virtual,iglesias2026conceptual}. From a philosophy-of-science perspective, this limitation also echoes Popper's critique of holism: because scientific descriptions are necessarily selective, no model can exhaustively represent a complex system in its totality \citep{popper1957poverty}. 

In contrast, \textit{bottom-up} approaches begin with available observations and use statistical and data-driven methods to identify relevant structures, associations, and recurring patterns that can then inform or complement existing knowledge of ecosystem functioning \citep{blair2021digital,campana2021role}. Supervised, unsupervised, probabilistic, and hybrid models can integrate heterogeneous data streams, identify anomalous combinations of stressors, reconstruct missing information, estimate species distributions, and project environmental conditions under alternative scenarios. Data-driven components can also be combined with physical and biological models, allowing empirical patterns to be constrained by established knowledge of ecosystem processes. This modular approach supports incremental development: the DTE can evolve as new observations, models, and computational services become available, without requiring a complete representation of the ecosystem from the outset. However, the data-driven approach also introduces important limitations. Statistical associations do not necessarily identify causal mechanisms, and model outputs may reflect biases, gaps, or scale mismatches in the source data. Predictions can become unreliable under environmental conditions that differ from those represented during model development. Furthermore, models trained for one ecosystem may not transfer directly to another, while opaque learning algorithms can make uncertainty and failure conditions difficult to communicate. A data-driven DTE must therefore preserve the provenance and processing history of its inputs, explicitly report data and model uncertainty, and distinguish evidence supported by quantitative analysis from plausible but unverified ecological interpretation. Principles of data findability, accessibility, interoperability, and reusability (FAIR) are particularly important in this context because they enable the DTE to be inspected, reproduced, extended, and connected to other environmental research infrastructures \citep{rebelo2023role,peters2024fair,islam2026fair,rashidi2026fair,vannini2026multi}.

A general issue with DTEs is that, even when data and models have been successfully integrated, their outputs remain difficult for heterogeneous users to access \citep{mihai2022digital,wu2023comprehensive,wang2023survey}. Environmental models generally expose their results through maps, technical reports, dashboards, or specialised software interfaces. These formats are useful to experts but can limit the ability of public authorities and citizens to formulate questions across disciplinary or modelling boundaries. Large language models (LLMs) offer a natural-language interface through which users can interrogate environmental information, request comparisons between scenarios, and receive explanations adapted to their needs. Nevertheless, an unconstrained LLM may generate convincing answers that are quantitatively inaccurate, unsupported by the available evidence, or based on generic knowledge rather than local data. This risk becomes particularly important when a conversational system operates close to environmental decision-making \citep{charitonidou2022urban,helbing2023digital,rahimi2024decide}.

%Two broad architectural strategies can be adopted to connect LLMs to a DTE. In an \textit{agentic} architecture, the LLM acts as the system’s central decision-making component: it interprets the user's request, autonomously retrieves information, selects and invokes available tools, plans and executes analytical operations, reasons over the resulting evidence, and constructs the final response \citep{NISA202669}. This architecture is flexible and can produce rich scientific narratives, but it delegates a substantial part of the analytical process to a probabilistic linguistic model. Consequently, it may provide limited quantitative grounding, invoke inappropriate information, or introduce unsupported relationships \citep{ding2026groundedscalingagenticai}. Conversely, in a \textit{reactive} architecture, question classification, data retrieval, statistical analysis, model execution, and scenario simulation are performed by specialised external components \citep{NISA202669}. The LLM receives the validated outputs of these components and primarily converts them into a clear, context-sensitive response. The separation between the data processing components and the LLM can improve operational control, process transparency, and traceability, although predefined analytical workflows may reduce response flexibility \citep{gabauer2026llm}, thereby limiting a model's ability to connect results to a broader ecological explanation.

Conversational ecosystem-modelling systems can allocate analytical responsibilities in different ways. In a predefined workflow, application logic determines which operations are available and how they are executed, while an LLM may extract query parameters and communicate the results. Alternatively, a system can rely primarily on the LLM to interpret retrieved ecosystem information and previously generated model outputs, identify relevant relationships, and formulate an answer. These design choices can influence implementation flexibility, scientific narrative, quantitative grounding, and answer reliability. The term \textit{reactive} architecture is used here to describe a workflow architecture with semantic routing and predefined analytical modules, consistent with the description of reactive agents as responding to current inputs through predefined rules or heuristics \citep{NISA202669}. The term \textit{agentic} architecture is used more broadly in this study to denote an LLM-led conversational architecture, in which the language model is tasked with interpreting the available evidence, formulating hypotheses and projections when needed, and constructing the analytical narrative. This distinction thus concerns the use of explicit, query-specific analytical workflows versus reliance on LLM-generated interpretations of retrieved evidence, rather than a universally exclusive classification of AI as reactive or agentic. 

This study investigates how these two strategies can support the implementation of a conversational DTE. It presents a general \textit{reactive}-DTE architecture for transforming heterogeneous environmental observations into harmonised spatial information, connecting them to biodiversity and ecosystem-risk models, and making their outputs accessible through a natural-language interface. Moreover, it compares this architecture to alternative \textit{agentic} DTEs to highlight differences in answer quality and completeness. Both architectures are based on FAIR and spatio-temporally harmonised data sources as well as on live species and environmental observations. The \textit{reactive} DTE integrates species distribution models and projections, multivariate ecosystem risk assessment, ecosystem scenario simulation, contextual information retrieval, and conversational response generation. These components form a modular evidence and modelling architecture in which scientific calculations remain distinguishable from linguistic interpretation.

The study compares two \textit{agentic} DTEs - based on (Meta) Llama 3.2 3B and (Google) Gemini 3.5 Flash LLMs, respectively - and four \textit{reactive} configurations - using Llama 3.2 3B, Gemini 3.5 Flash, Llama 4 Scout, and Qwen3 8B LLMs, respectively. For the benchmarking, the DTEs are adapted to simulate the Massaciuccoli Lake basin in Tuscany, Italy. The basin is a 114 km² agro-wetland and ecotonal system in which lake, marshland, terrestrial, agricultural, urban, and coastal influences interact. Its biodiversity and hydrological conditions are affected by climate variability, water availability, vegetation dynamics, land-use change, infrastructure, and intensive agricultural activities, which occupy approximately 40\% of the area. Therefore, the basin provides a suitable test environment for investigating how multiple pressures interact and how their effects can be communicated to monitoring authorities and citizens. 

%The Massaciuccoli DTEs integrate satellite and \textit{in situ} observations, geospatial layers, citizen science records, and species observations, harmonised on a common spatial grid \citep{vannini2025fair}. In the \textit{reactive} DTEs, unsupervised species distribution models \citep{coro2024climate,bove2026estimating} reconstruct historical patterns and project habitat suitability. Their outputs are combined with environmental and anthropogenic variables in a multivariate, unsupervised ecosystem risk assessment model \citep{vannini2026multi} that identifies conditions associated with habitat degradation and biodiversity loss. The DTEs are tasked with synthesising the knowledge base formed by the data, species distributions and projections, and the ecosystem risk assessment to provide a consistent answer to the user. 

%They are evaluated using questions that cover background data availability, correlations between environmental and ecological variables, model-based variable importance, changes in ecosystem risk under simulated perturbations, comparisons between management scenarios, species distribution and habitat structure, and general questions about ecosystem conditions.

The DTEs are assigned the role of responding as a scientist would. They enable users to ask which variables are available, which factors characterise high-risk areas, how an individual species' habitat is structured, how risk could change under a combined climatic scenario, or which of two interventions is expected to be more effective in reducing ecosystem risks. Their answers are evaluated according to correctness, scientific narrative, quantitative grounding, ecological interpretation, hallucination control, use of local ecological knowledge, and linguistic fluency. 

%The main contribution of this study is the proposal of a reactive DTE architecture and an evaluation of how different architecture-LLM combinations affect response quality, while exploring how the allocation of scientific reasoning between the language model and external computational components affects reliability. The results provide evidence that \textit{reactive} architectures can improve correctness, quantitative fidelity, and control of hallucinations when communicating environmental-model outputs, while \textit{agentic} architectures retain advantages in scientific narrative, contextual interpretation, and ecological storytelling. This trade-off identifies the design requirements for future DTEs that must combine computational reliability with accessible and meaningful communication.

The main contribution of this study is the implementation and evaluation of a modular conversational DTE framework that integrates harmonised geospatial data, periodically updated observations, scientific documents, and specialised analytical models. Its reactive architecture uses semantic routing to select predefined analytical workflows and separates numerical evidence production from language generation. The general routing-and-analysis pattern is an established architectural approach; the contribution lies in its integration and evaluation for ecosystem exploration and risk-related questions. The comparison with agentic baselines, as operationally defined above, examines differences in answer quality between explicit analytical workflows and LLM-led interpretation of ecosystem evidence.

The paper is structured as follows: Section \ref{sec:overview} reports an overview of DTE designs and implementations. Section \ref{sec:methods} describes the components of the proposed \textit{reactive} DTE architecture and specifies how it was implemented in the Massaciuccoli Lake case. It also includes the specification of the evaluation strategy and the baseline \textit{agentic} DTEs used for performance comparison. Section \ref{sec:results} reports the results of the DTE evaluation, and Section \ref{sec:discussion} discusses the results, clarifies limitations, and draws conclusions.

\section{Overview}
\label{sec:overview}
The term \textit{digital twin} refers to systems with different architectures and capabilities. In its strictest sense, a DT is a virtual representation that remains connected to its physical counterpart via automatically updated data flows and can exchange information in both directions \citep{kritzinger2018digital}. This distinguishes it from a \textit{digital model}, which has no automatic connection to the physical system, and a \textit{digital shadow}, in which information flows automatically only from the physical system to its digital representation. 

This definition cannot be transferred straightforwardly from engineered assets to natural ecosystems, because ecosystems cannot be comprehensively instrumented or controlled. DTEs are more broadly understood as dynamic representations that combine observations, models, and computational services to monitor an environmental system and investigate its possible responses to natural and anthropogenic changes \citep{blair2021digital}. They are intangible objects whose parameters cannot be easily altered without affecting the entire ecosystem.

DTEs can be placed along a continuum according to the relative importance of their data, modelling, analytical, and interaction layers. At one end, infrastructure-oriented systems organise and provide access to distributed data and computational resources (data-oriented DTEs) \citep{damasevicius2024,MAJIDINEZHAD2024114065}. At the other end, model-oriented systems add numerical simulations to reconstruct environmental conditions and explore hypothetical scenarios \citep{lin2026digital}. Among these systems, those based on artificial intelligence (AI) use machine learning to infer relationships from heterogeneous observations, integrate specialised models, detect anomalous conditions, and provide user-oriented access to the underlying evidence \citep{skakala2023,islam2026fair}. In-between, hybrid infrastructure-based AI systems balance the advantages of both system types.

Data-oriented DTEs address the fragmentation of environmental observations across satellite archives, \textit{in situ} monitoring networks, research infrastructures, governmental repositories, and citizen science platforms \citep{islam2026fair}. Their main functions include data discovery, standardised access, harmonisation, provenance tracking, scalable processing, and machine-to-machine interoperability. The European Digital Twin Ocean infrastructure \citep{edito2026}, EDITO, is a major example. It provides a shared environment in which ocean data, models, processing tools, and applications can be accessed and combined, building on resources such as Copernicus Marine Service and EMODnet \citep{copernicus_marine,emodnet_general}. EDITO is an enabling infrastructure for developing multiple DT applications rather than being a single comprehensive ocean simulation. 

The United Nations' Ocean Decade programme Digital Twins of the Ocean \citep{ditto_whitepaper_2023}, DITTO, adopts a broader definition of DTE. It describes ocean DTs (DTOs) as virtual representations linked to the real ocean via observations and capable of exploring hypothetical scenarios using models. Its framework includes observing systems, frequently updated repositories, analytics and prediction engines, computing resources, interactive interfaces, and capacity development \citep{bahurel2023ditto}. DITTO is thus a methodological and coordination framework in which data infrastructures constitute one component rather than the complete DT.

A common feature of DT infrastructures is the reliance on 
FAIR data to reduce the effort needed to locate and preprocess heterogeneous information and reuse data and models across applications \citep{rebelo2023role,peters2024fair,islam2026fair,rashidi2026fair}. However, technical interoperability does not guarantee an integrated, holistic understanding of ecosystem behaviour. Data may remain semantically disconnected even when they share the same format, and visual superposition may be mistaken for functional integration. Without validated models, uncertainty propagation, and scenario-generation capabilities, a data-oriented system remains closer to an advanced data platform or digital shadow than to a complete DTE.

In contrast, model-oriented DTs place numerical simulation at the centre of the architecture. They assimilate observations into mechanistic, process-based, or statistical models to reconstruct system states, produce forecasts, and evaluate \textit{what-if} scenarios \citep{Partridge2026}. Marine implementations may combine hydrodynamic, biogeochemical, sediment, habitat, food-web, and socioeconomic models. For example, Destination Earth (DestinE) \citep{destine_initiative} - a European Union initiative for developing high-precision digital representations of the Earth system - includes a Climate Change Adaptation digital twin that combines observations with coupled atmosphere-ocean models to produce high-resolution climate projections and sector-specific impact information. It supports multidecadal simulations and storyline experiments in which historical extreme events are reproduced under different climatic conditions. Similarly, the ILIAD project \citep{papachristopoulou2025digital,iliad_project} combines interoperable ocean data, modelling services, analytics, and interactive interfaces in multiple thematic applications. Maritime DTs developed by NATO Science and Technology Organisation and the Centre for Maritime Research and Experimentation \citep{nato_sto_website,nato_cmre_website} also connect real and simulated assets, environmental models, sensor simulators, and operator services to support mission planning, training, interoperability testing, and system validation. 

The principal advantages of these systems are their physical consistency, explicit representation of known processes, and ability to generate controlled scenarios. Their development, however, is demanding, and implementations for specific areas are not readily reusable across different areas. Models produced by different communities may use incompatible spatial scales, variables, assumptions, and software environments. Calibration requires observations that often are sparse, while high-resolution simulations impose considerable computational and storage costs. Furthermore, even sophisticated physical models may only partially represent biodiversity, human behaviour, governance, and ecosystem services. 

In this context, data-driven, AI-based DTEs can automatically identify variables associated with ecosystem status and environmental risk by incrementally integrating and connecting data streams and specialised models. Their tendency towards a holistic character emerges from the relationships established among these components as new evidence and modelling capabilities become available. Specialised AI models process the data streams, while decision-support and optimisation components evaluate strategies intended to minimise ecosystem impacts. However, only a few proposals for these systems are currently available, with limited actual implementations \citep{campana2021role,Campana2025Holistic}. A main limitation is that statistical associations do not establish causality, predictions may fail outside the conditions represented in the training data, and model outputs can inherit observational biases.

The interface of a DTE towards its users and stakeholders can build on established work in scientific question answering, retrieval augmentation, and tool-assisted computation. The value of information retrieval from authoritative documents and accurate question answering about climate change via an LLM to support decision making has been demonstrated in both automatic systems, such as ChatClimate \citep{vaghefi2023chatclimate}, ClimateGPT \citep{thulke2024climategpt} and similar tools \citep{juhasz2024responsible}, and semi-automatic systems, such as ClimaQA \citep{manivannan2025climaqa}. Beyond documentary retrieval, LLMs have been integrated with GIS tools (e.g., GeoGPT, \cite{zhang2024geogpt}) to efficiently explore and query geospatial datasets, sometimes enabled with specialised agents for spatial data retrieval and analysis (e.g., GeoAgentic-RAG, \cite{liang2026geoagentic} and ClimSight \cite{kuznetsov2025transforming}). In the broader context of environmental informatics, conversational and tool-routed architectures have increasingly sought to bridge unstructured natural-language requests with structured computational workflows. Autonomous multi-agent systems and tool-augmented frameworks (e.g., GeoAgentBench, \cite{yu2026geoagentbench}, AQUAH \cite{yan2025aquah}, and Earth-Agent \cite{feng2025cognitive}) leverage LLMs to plan and dynamically invoke APIs, yet they generally encounter challenges with explainability, error propagation, and hallucinated control trajectories when applied to complex, data-intensive natural environments \citep{munir2026agentic}.

The present framework similarly connects language generation to external evidence and computation, but adopts embedding-based routing to predefined ecosystem workflows rather than LLM-directed tool planning. This choice prioritises inspectable analytical execution for recurring questions about habitat, stressors, and scenarios. Its contribution lies in integrating these capabilities within an ecosystem-specific DTE and evaluating complete configurations against LLM-led retrieval baselines.

The DTE interpretation adopted in this study builds on the integration of environmental observations and scientific modelling discussed by \cite{blair2021digital}, the ecological model–data perspective of \cite{de2023digital}, and the integrated ecosystem representation proposed by \cite{Campana2025Holistic}. The term DTE here denotes a spatially explicit, periodically updated virtual representation of an identified ecosystem that combines observations, scientific models, and interactive analytical services. This operational interpretation builds on the ecological model–data integration perspective discussed above. In the proposed implementation, the physical counterpart is the Massaciuccoli Lake basin; harmonised geospatial variables provide its baseline representation; periodically refreshed observations describe recent conditions; and executable habitat and risk-scenario workflows support conditional exploration. The conversational component provides access to these resources and operations and represents the interaction layer of the DTE.

\section{Methods}
\label{sec:methods}
\begin{figure}
    \centering
    \includegraphics[width=1\linewidth]{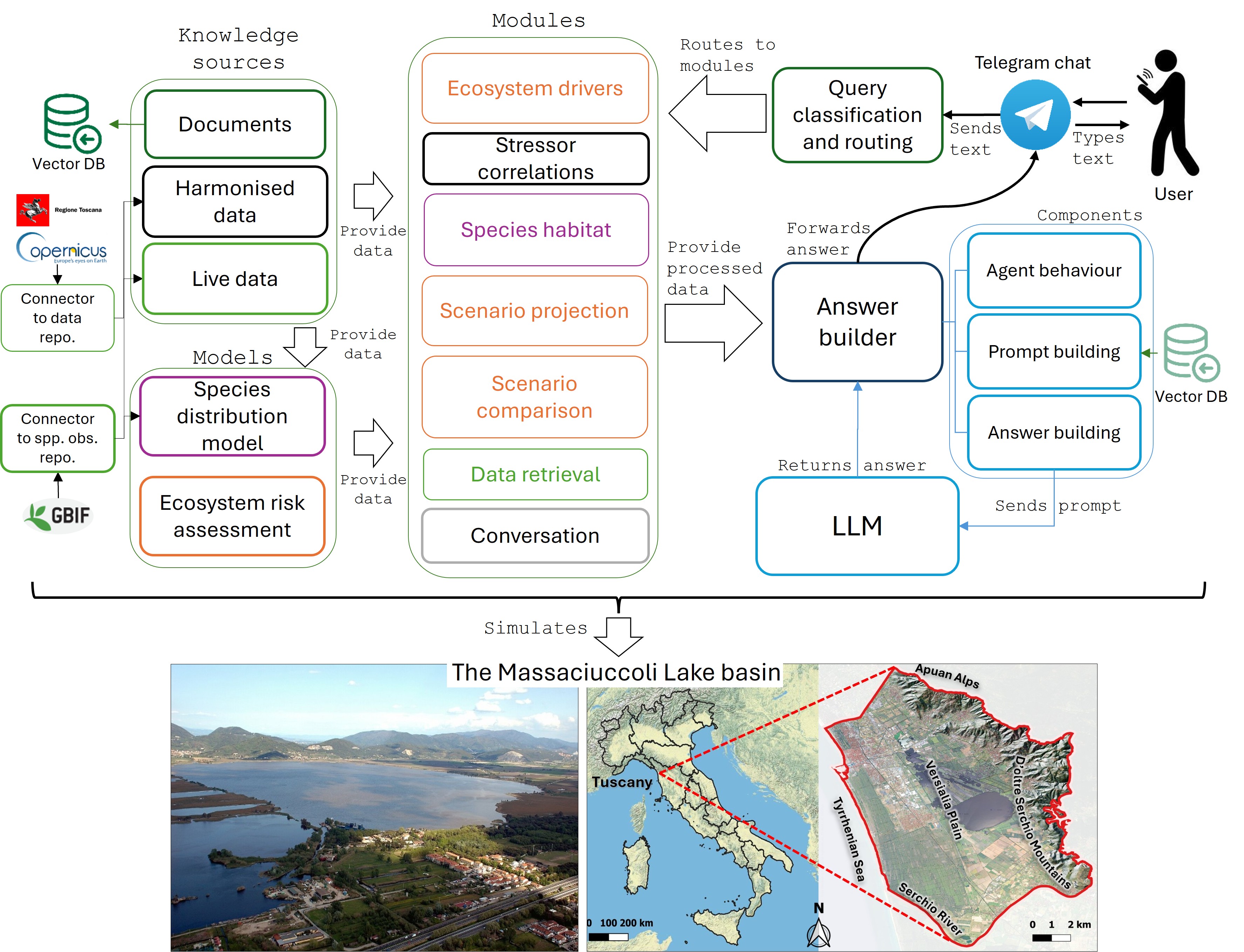}
    \caption{Architectural schema of the proposed \textit{reactive} agent-based digital twin of a natural ecosystem. Note that the \textit{vector DB} icon refers to the same vector database hosting the embedded documents.}
    \label{fig:reactive_agent}
\end{figure}

This section outlines the components of the reactive DTE architecture proposed in this work, following the schema presented in Figure \ref{fig:reactive_agent}. In the proposed reactive architecture, each incoming query activates an existing workflow selected through semantic routing. Where required, the LLM extracts query parameters before module execution and subsequently formulates the answer from the resulting evidence. It does not autonomously construct or revise the analytical execution sequence. The term \textit{reactive} thus concerns workflow control, while individual components employ machine learning, statistical inference, and generative language models. The subsequent sections detail the overall design of each component and clarify their adaptation to the Massaciuccoli Lake basin case study. The evaluation methodology concludes this section by examining the effectiveness and limitations of this approach compared with an alternative agentic-AI architecture. 

\subsection{Knowledge sources}
\label{sec:knowledgesources}
Key sources of the proposed DTE architecture can be categorised as: (i) documents describing background knowledge, studies and evaluations of the components and ecosystem status and services; (ii) background data, expressed as a harmonised collection of spatio-temporal data representing slowly changing environmental, biotic, anthropogenic, and climatic features; and (iii) live data, reporting information from active sensor networks residing in the analysed area. The next sections describe each of these source types, represented by the \texttt{knowledge sources} block of Figure \ref{fig:reactive_agent}.

\subsubsection{Documents}
\label{sec:documents}
The DTE background knowledge base is constructed from scientific publications concerning the ecosystem under study. The document collection should cover the ecosystem's biotic and abiotic characteristics, ecosystem services, anthropogenic pressures, climate-change effects, land-use patterns, human activities, and urbanisation. These publications should provide relatively stable scientific and contextual knowledge that complements both slowly changing geospatial datasets of ecosystem variables and live observations acquired from sensors or external services. The literature should support the interpretation of structured and sensor data describing measured environmental conditions by providing information on local ecological processes, historical changes, known pressures, and previously observed relationships. The documents thus constitute contextual sources that complement the quantitative evidence available to the DTE.

To enable semantic retrieval, the documents are indexed in a vector database. Each document is first converted into plain text, cleaned of formatting artefacts, and divided into smaller segments (\textit{chunks}). Cleaning is conducted by removing page artefacts and non-scientific editorial material, repairing line breaks and hyphenation, and normalising white-space and abbreviations. Chunking is necessary to limit the amount of unrelated information included in each retrieval unit while preserving sufficient context for interpretation. The selected chunk size must balance two competing requirements: chunks that are too short may fragment a scientific argument, whereas excessively long chunks may contain several unrelated concepts and reduce retrieval precision. Whenever possible, chunk boundaries should be aligned with sentences or paragraphs to preserve semantic coherence, while metadata should maintain the connection between each chunk and its source document.

For the Massaciuccoli Lake case study, ChromaDB \citep{chromadb} was used to create a dedicated vector collection named \texttt{documents}. Extracted text was normalised by collapsing whitespace and divided into consecutive, non-overlapping chunks of up to 1,200 characters. Each chunk was transformed into a dense numerical vector using \texttt{Nomic}, a text-embedding model designed for semantic search and document retrieval  \citep{nussbaum2024nomic}. This model maps passages with similar meanings to nearby positions in a multidimensional vector space. At query time, the same embedding model represented the user question, and ChromaDB returned the six nearest chunks. Candidates were retained when their squared Euclidean distance was no greater than 0.6, with duplicate passages removed and the remaining passages ordered by increasing distance.

The documents comprised 12 research articles in English and covered local- and regional-scale studies covering the evolution of territorial aspects over 20 years, future projections of biodiversity and ecosystem services in the Massaciuccoli basin, as well as known societal, industrial, and service problems of the basin and its surrounding municipalities \citep{vannini2025fair,viciani2017natura,coro2024climate,pasquetti2024hydrodynamics,venturi2021exploring,da2020coastal,rossetto2010water,bertacchi2015vegetation,ciccolini2013restoration,pistocchi2012simple,rossetto2010surface,baneschi2007geochemical}.  

Document fragment retrieval forms the basis of a Retrieval-Augmented Generation (RAG) process in the DTE. The agent adds the most relevant chunks as complementary contextual information to the final answer (Section \ref{sec:prompt_builder}). This enables the agent to relate numerical results to the documented ecological and territorial context, producing more informative answers about ecosystem processes, pressures, and potential responses. Maintaining a distinction between retrieved background knowledge and model-derived evidence also improves traceability and reduces the risk that general information from the literature is presented as a direct result of the DTE.

\subsubsection{Harmonised data}
\label{sec:harmonised_data}

The DTE relies on harmonised geospatial variables suitable for ecosystem risk assessment (\textit{ecosystem dataset}). Spatial and temporal harmonisation is essential for constructing this dataset, as comparisons and multivariate analyses require all variables to be aligned across locations, resolutions, and time periods. The equirectangular projection under the WGS 84/EPSG:4326 geographic coordinate reference system should be preferred for the data to ensure a constant angular cell size between adjacent cells. FAIRness is equally important for preserving data provenance, accessibility, interoperability, and potential reuse. Together, these properties transform heterogeneous geospatial resources into a machine-processable evidence base. Such preparation is a prerequisite for many DTEs, which depend on consistently aligned and traceable data to connect observations, models, simulations, and user-facing services (Section \ref{sec:overview}). 

The ecosystem dataset should represent slowly changing conditions, in contrast to observations from live sensors and external services. Its function should be to provide a spatially complete and temporally comparable representation of the background conditions against which newly acquired observations and simulated changes could be interpreted. 

Data selection follows a general-purpose, semi-automatic approach named Geodata Layers Harmonization Methodology (GLHM) \citep{vannini2025fair}, a general workflow organised into four phases: geodata census, selection, alignment, and publication. Candidate datasets are first identified and documented with their spatial, temporal, descriptive, and provenance metadata. Then, the workflow automatically estimates the FAIRness degree of each accessible dataset. This is done by assigning a score of \(0\), \(0.5\), or \(1\) to each FAIR sub-principle, depending on whether its requirements are unmet, partially met, or fully met, and summing these scores to obtain an overall FAIRness index. Datasets are classified as having insufficient, sufficient, or good FAIRness; they are retained only if the essential ``findability'' requirement (easy to locate on a repository) is met at least partially and the data can be downloaded directly. 

The variables selected are semantically standardised, assigned to common time frames, reprojected into a common coordinate system, clipped to the study area, and resampled to a unified spatial resolution. All variables are aligned to a common spatial grid and converted to numerical form. Proportional categorical variables are mapped to a normalised scale between 0 and 1, while presence-absence variables are encoded as Boolean 0/1 values.
%The numerical stressors are \(z\)-standardised to yield a mean of zero and a standard deviation of one, preventing variables with larger measurement ranges from dominating the multivariate analyses. If insufficient ecological evidence supports variable-specific weighting, no additional variable-specific weights should be used. 

When multiple datasets represent the same variable and temporal frame, preference is given to the dataset closest to the latest representative year, followed by the one with the highest FAIRness score, and then the one with the longest temporal coverage. When spatial interpolation is needed, it is performed using the Bayesian Interpolation Model with Advection-diffusion Constraint (BIMAC), a non-parametric, machine-learning-based model that is case-study-independent and adaptable from static to fluid-dynamic scenarios \citep{coro2024open}. 

These techniques make the construction of the ecosystem dataset independent of the specific application case and reusable for building new DTEs. As a demonstration, previous work has used the GLHM workflow for marine ecosystems at global, Mediterranean, and basin (Baltic Sea) scales \citep{coro2023global,pavirani2025assessing,pavirani2025ecosystem,Pavirani2026}. 

The selected ecosystem variables are then reprocessed to represent ecological \textit{stressors}, namely spatio-temporal conditions whose high values may contribute to changes in ecosystem status and, potentially, to critical ecological transitions. The stressors should be separated into \textit{harmful forces}, representing climatic or anthropogenic drivers that can negatively affect habitats (e.g., increases in land use, tree cover and grassland loss, imperviousness, temperature and aridity, and decreases in precipitation), and \textit{fragile conditions}, representing ecosystem properties associated with sensitivity or limited resilience (e.g., plant productivity, biodiversity richness, and the presence of trees and grasslands). A stressor can correspond directly to a geospatial variable, such as increasing land imperviousness, or to its inverse when low values represent adverse conditions, as in the case of decreasing precipitation. 

For the Massaciuccoli Lake basin, ecosystem variables were selected from the publicly available Massaciuccoli cartographic repository and its accompanying geographic Atlas \citep{zenodorep1,zenodorep2,vannini2025fair}. The GLHM workflow identified 75 geospatial variables distributed across 148 raster and vector datasets comprising environmental, climatic, pedological, biological, and anthropogenic information. All data were represented at a common spatial resolution of \(0.0005^\circ \times 0.0005^\circ\), over an equirectangular projection, corresponding to approximately \(50 \times 50\) m, for a total of 49,382 georeferenced cells covering the basin. The available information spanned five temporal snapshots, from the remote past to projections for 2100 under different Representative Concentration Pathways. From this broader collection, the following 12 stressors were extracted for the 2015-2025 decade: 
\begin{enumerate}
    \item Current land use: Land cover and use from the Corine Land Cover European inventory \citep{buttner2014corine}, with 0 indicating no use and 1 land used for human activities.
    \item Land use increase: Change in land use and land cover, where 1 indicates territories recently altered by human activities (such as mining, urban development, military operations), 0.5 corresponds to areas used for agriculture, and 0 represents unchanged land and other natural types like woods, wetlands, and water bodies.
    \item Land imperviousness increase: Average per cent sealing density change.
    \item Tree cover density: Average per cent tree cover density.
    \item Tree cover loss: Loss or unchanged tree cover per cell (1 and 0 values, respectively).
    \item Grassland presence: Presence or absence of grassland per cell (1 and 0 values, respectively).
    \item Grassland loss:  Loss or unchanged grassland cover per cell (1 and 0 values, respectively).    
    \item Plant productivity: Average daily plant phenology index measured as days of vegetation growth.
    \item Temperature increase: Average annual temperature increase (in °C) of the air at 2 metres above the surface minus the average over 1971-2010.
    \item Atmospheric aridity increase: Average per cent amount of potential evapotranspiration (evaporation that would occur if sufficient water provisioning were available from the water sources) minus the average over 1971-2010.
    \item Precipitation decrease: Average per cent water that falls as rain, snow, sleet, or hail minus the average over 1971-2010.
    \item Animal biodiversity: Annual species richness index representing the total number of birds, fish, crustaceans, insects, clams, mammals, reptiles, amphibians and other animals potentially present per cell.
\end{enumerate}

In previous work, these features demonstrated potential in reliably assessing current ecosystem risks in the Massaciuccoli basin associated with forest and grassland degradation and the consequent loss of biodiversity \citep{vannini2026multi}.

The DTE loads the ecosystem dataset into memory as a table. Its columns correspond to the stressor variables, while the georeferenced records represent the spatial cells. The first two rows provide the metadata required by the agent modules: the first contains the name of each stressor, and the second its corresponding extended description. These descriptions are also indexed in the RAG knowledge base, where they provide contextual information about the variables' meanings and support the identification of their semantic connections and potential interrelationships. They can therefore be retrieved alongside the scientific-document fragments to help the agent interpret analytical results and relate the variables to the broader ecological context. The subsequent rows contain the numerical values used for later processing by the models and modules (Sections \ref{sec:models} and \ref{sec:modules}).

\subsubsection{Live data}
\label{sec:live_data}
The slow-changing variables in the \textit{ecosystem dataset} should be complemented by regularly updated observations that reflect recent conditions in the study area. These observations are consolidated into an in-memory table named the \textit{live dataset}, which provides the agent with a compact representation of the latest climatic, ecological, and land-cover information. 

In this context, \textit{live} indicates that the data are periodically refreshed and aggregated from their original sources, rather than necessarily acquired and processed in real time. Selecting an appropriate temporal aggregation interval is essential. High-frequency sensor measurements may generate unnecessarily large datasets, whereas excessive aggregation may hide relevant variations \citep{coro2018enhancing,scarponi2018collection}. The aggregation time interval should therefore reflect both the production rate of each source and the information required by the DTE. 

Different sources may be updated at substantially different frequencies: weather stations can produce many observations per day, land-cover products are updated less frequently, and species-observation repositories depend on the irregular submission and publication of occurrence records. These differences must be considered when constructing the live dataset to avoid treating data availability or sampling intensity as equivalent across variables. Live data can originate from public monitoring networks, such as Copernicus \citep{copernicus_marine}, or restricted sensor infrastructures for which the agent has been granted access \citep{scarponi2018collection,vannini2025fair}. They can be retrieved from repositories and services that periodically publish new observations or those generating continuous sensor streams. For example, the Global Biodiversity Information Facility (GBIF) \citep{gbif_infrastructure} makes species-occurrence records available as they are contributed by research institutions, monitoring programmes, and citizen-science initiatives.

For the Massaciuccoli case study, climatic measurements were obtained from the Servizio Idrologico Regionale della Toscana (SIR) \citep{sir_toscana_portal}, an open monitoring network providing meteorological and hydrological observations. Data about land use, impervious surfaces, tree cover, and grassland were derived from GEOscopio, the geographical information service maintained by Regione Toscana \citep{geoscopio_toscana}. Species occurrence data were obtained via spatial queries to GBIF to retrieve occurrence records for species reported within the Massaciuccoli Lake basin. These records were spatially aggregated at approximately 50 m resolution to estimate species richness, expressed as the number of distinct species observed within each grid cell. 

Since the DTE was intended to provide general information and perspectives on ecosystem status rather than short-term meteorological alerts, the live dataset used annual aggregation. The weather-derived variables were: average temperature change during the latest year; average change in evaporation relative to the 1971-2010 reference average; and average rainfall-derived water availability, expressed as a percentage of the 1971-2010 reference average. Moreover, the biodiversity variable reported the average number of animal species per spatial cell. Other variables described land imperviousness, tree-cover density, grassland cover, and land used for human activities, all expressed as percentages. For each climatic, biodiversity, and land-cover variable, the dataset reported its change relative to the preceding year. The live dataset is updated annually, enabling the agent to describe both the latest aggregated conditions and their recent direction of change. 

Live data acquisition and preparation in a DTE should use caching mechanisms and rely upon high-availability e-infrastructures. For example, the Massaciuccoli DTE retrieved live data from the CNR central service infrastructure \citep{cnr_centro_servizi_portal,cnr_portale_servizi_2021}, a high-availability research infrastructure that serves as a data aggregation service, relaying data from sensors and services to the agent. The source data and processed products resided in the infrastructure's distributed storage system. An automated workflow running on the infrastructure daily retrieves observations from the SIR network and the public repositories of  Regione Toscana, processes them, and updates their daily aggregates. This workflow also retrieves species occurrence records directly from GBIF to generate aggregated biodiversity reports. 

At the end of each update cycle, the resulting aggregate replaces the previous file as a new version in the infrastructure storage system. To minimise unnecessary data transfers and processing, the DTE should check whether the remotely stored file has changed before downloading it. The live dataset is then reloaded as an in-memory table. The update mechanism in the proposed DTE implementation is independent of the specific application case and is implemented as a standalone module that provides the agent with up-to-date information, with an update frequency determined by the application requirements. In a new application, the module internally requires changes to the connectors for the sensor and data networks.

\subsection{Models}
\label{sec:models}

This section describes the auxiliary models used by the proposed DTE, represented in the \texttt{models} block of Figure \ref{fig:reactive_agent}. These models provide pre-processed data to the agent modules (Section \ref{sec:modules}) for further processing.

\subsubsection{Species distribution model}
\label{sec:sdm}

Species distribution models (SDMs) estimate the spatial distribution of a species by relating its observed occurrences to environmental, climatic, and anthropogenic conditions measured across a study area \citep{elith2009species}. Once this relationship has been estimated, the model can assign a species-presence probability to every spatial cell in the study area. Repeating the projection with environmental conditions representing different periods or scenarios also enables the analysis of potential changes in species presence over time.

The distinction between an ecological niche model (ENM) and an SDM is relevant in this context \citep{peterson2012species}. An ENM estimates the environmental conditions compatible with a species' persistence and projects these conditions across geographic space to indicate \textit{potential habitat suitability}. An SDM aims to approximate the species' \textit{realised spatial distribution} and may additionally account for accessibility, dispersal, biotic interactions, historical constraints, or known limits to its native range. In practice, the terms are frequently used interchangeably, particularly for correlative presence-only models, but have significant complementarities. The model described here is operationally referred to as an SDM because it generates and analyses a spatial distribution of species. However, its core uses an ENM-derived potential habitat distribution based on estimated functional relationships between occurrences and environmental conditions, which is ultimately transformed into the species' realised distribution through spatial re-analysis.

The output of an SDM can be processed to estimate the proportion of a region that contains suitable habitat and species presence, and assess whether this region forms a continuous area or several isolated patches. This analysis can estimate distances among patches and assess whether a dominant core is surrounded by smaller satellite areas, providing landscape-level indications of habitat spread, fragmentation, and connectivity. The analysis can also identify the environmental or anthropogenic predictors that contribute most strongly to the estimated distribution, thereby indicating the conditions to which the species' presence is most sensitive.

The proposed DTE embeds the open species-distribution workflow presented by \citet{coro2025open}, which supports four complementary ecological niche models: AquaMaps, Maximum Entropy, Artificial Neural Networks, and Support Vector Machines. The complete workflow can automatically optimise the models, convert continuous habitat suitability into species presence distributions, combine individual projections into an ensemble, and aggregate multiple species distributions into a species-richness index. 

For on-demand use within the DTE, the agent employs only the Maximum Entropy (MaxEnt) model. MaxEnt is a presence-background ENM that estimates the probability

$$
\pi(\mathbf{x}) = P(\text{suitable habitat}\mid\mathbf{x}).
$$

where \(\mathbf{x}\) is the vector of environmental and anthropogenic conditions associated with a spatial cell. The model searches for a probability distribution with maximum entropy - that is, the least informative distribution consistent with the available constraints - subject to constraints derived from the environmental conditions observed at species-presence locations. Background points sampled from the study area characterise the environmental conditions available to the species and provide a reference against which the presence locations are distinguished. Consequently, MaxEnt can be trained when reliable absence observations are unavailable, a common condition in biodiversity repositories and for opportunistic species observations. The SDM applies a spatial correction algorithm to constrain potential suitability based on the observed species' occurrence distribution, thereby transforming the ENM projection into a closer approximation of a realised species distribution. Technical details about this process are provided in the related paper \citep{coro2025open}.

MaxEnt was selected among the supported models because it offers a suitable compromise among execution speed, limited parameterisation, compatibility with presence-only observations, and interpretability. It can train and project a distribution quickly enough to manage a species selected at query time. Moreover, it reports the relative contribution of each predictor variable and the Area Under the Receiver Operating Characteristic Curve (AUC). Predictor contributions indicate which variables most strongly influenced the fitted suitability function, while the AUC measures the model's ability to discriminate presence locations from background locations across alternative classification thresholds. A high AUC indicates strong discrimination within the evaluation data and can serve as an approximate indicator of model quality when many species observations are available.

%The use of a single presence-only model is consistent with the purpose of the DTE, which is to rapidly extract general, landscape-level characteristics of species distributions rather than produce a detailed conservation map. Indeed, local errors and individual model biases have progressively less influence when distributions are spatially and temporally aggregated to identify macroscopic biodiversity patterns \citep{weber2004scale,queiroz2021reply}. The SDM therefore supports ecosystem exploration and hypothesis generation rather than replacing detailed spatial species distribution estimation.

The use of a single presence-background model is consistent with the DTE's purpose of rapidly estimating landscape-level habitat characteristics. Occurrences are associated with the environmental grid, and projected distributions are summarised through basin-level coverage and patch descriptors. This aggregated representation can effectively reduce, although not eliminate, sensitivity to certain isolated positional discrepancies, particularly those within a modelling cell \citep{weber2004scale,queiroz2021reply}. MaxEnt further mitigates sampling biases by using background locations rather than confirmed absences. The resulting habitat descriptors therefore support ecosystem exploration and hypothesis generation, which can yield reliable, aggregated perspectives on species distributions (as particularly demonstrated for the Massaciuccoli Lake basin in \cite{coro2024climate}), whereas their use for site-specific management would require corroborating observations and expert interpretation.

\subsubsection{Ecosystem risk assessment model}
\label{sec:ecosystem_risk_assessment}

The term ``ecosystem risk'' refers to the likelihood that interacting environmental and anthropogenic pressures may cause an ecosystem to degrade, lose resilience, or transition towards a less favourable ecological state \citep{keith2013scientific,bland2017guidelines}. The DTE agent integrates the output of the ecosystem risk assessment (ERA) framework by \cite{pavirani2025assessing}. This framework models the competition of multiple stressors and the detection of nonlinear patterns or anomalies associated with potentially critical ecosystem conditions. As input, it requires aligned data on climatic, biotic, anthropogenic, and environmental variables, such as those in the \textit{ecosystem dataset}. The variables are transformed so that \textit{higher} values represent conditions of higher ecosystem stress (stressor variables). The numerical stressors are then \(z\)-standardised to yield a mean of zero and a standard deviation of one, preventing variables with larger measurement ranges from dominating the multivariate analyses. %If insufficient ecological evidence supports variable-specific weighting, no additional variable-specific weights should be used. 

The ERA framework's objective is to identify hotspot areas where \textit{harmful forces} interact with \textit{fragile conditions}. Risk conditions are inferred from multivariate, nonlinear combinations of the stressor values across the analysed area. First, a Variational Autoencoder (VAE) identifies anomalous stressor configurations. The VAE learns a latent representation of the stressor dataset and attempts to reconstruct the original observations from this representation. Stressor vectors that are poorly represented by the learned data structure are interpreted as anomalous and, consequently, as potentially associated with ecosystem risk \citep{pavirani2025assessing,bove2026estimating,vannini2026multi}. Second, Multi K-Means clustering is applied to organise the anomalous stressor vectors into internally homogeneous groups, with the number of clusters selected automatically. Third, statistical analysis identifies clusters characterised by unusually elevated stressor values for most variables (\textit{closer-attention} clusters). Specifically, values above the 75th percentile of the corresponding overall individual distributions are considered elevated. Additionally, among the closer-attention clusters, those with high values for biodiversity-related stressors (e.g., high animal biodiversity, tree cover density, and grassland presence) are retained. 

Finally, the retained clusters are assigned an ecosystem-risk class as follows: a \textit{high-risk} class is assigned when more than half of the stressor variables have elevated average values. A \textit{medium-risk} class is assigned when at least one stressor (apart from the biodiversity-related stressors) has an elevated average value. All other clusters, including those not retained, are classified as \textit{low risk}. 

This classification is finally assigned to each cell in the \textit{ecosystem dataset}, depending on the cluster to which its associated stressor vector belongs. It will serve as the basis for correlation analyses and what-if scenario simulations implemented by the DTE modules.

\subsection{Agent modules}
\label{sec:modules}

This section describes the modules of the proposed DTE agent, with the Massaciuccoli DTE as the principal example of implementation. In line with the general advice by DTE initiatives (Section \ref{sec:overview}), in its current design the agent can answer a minimal set of ecologically important query types, which include (i) the identification of the principal ecosystem drivers; (ii) the calculation of the correlations among the ecosystem stressors; (iii) the explanation of species' habitat composition and structure; (iv) projections of future ecosystem scenarios; (v) a comparison between future hypothetical ecosystem scenarios; and (vi) the retrieval of the currently available data. 

The DTE uses an LLM accessible as a service, whose IP address is specified in the DTE configuration file. Currently, Ollama \citep{marcondes2025using} and Google Gemini Enterprise Agent Platform \citep{google2026geminienterprise} are supported as model-hosting services.

The next sections describe the specific modules that handle the query types above (\texttt{modules} block in Figure \ref{fig:reactive_agent}), as well as the auxiliary modules for query classification and answer production.

\subsubsection{Query classification and routing}
\label{sec:query_routing}

The query-routing mechanism identifies the type of information requested by the user and activates the corresponding analytical module. This step is necessary because different questions require different data sources and processing operations. For example, a request about the principal drivers of ecosystem risk requires a variable-importance analysis, whereas a question about habitat fragmentation requires a species distribution model. Submitting both questions to the same general-purpose prompt directly would limit control over the evidence and calculations used to construct the answers.

The routing mechanism distinguishes the following seven query categories:

\begin{enumerate}
\item \textit{Ecosystem drivers}: questions concerning the variables that contribute most strongly to ecosystem state, stability, or risk. Examples include ``How does tree cover affect ecosystem stability?'', ``What factors drive ecosystem health?'', and ``How does biodiversity affect ecosystem-change risk?''

\item \textit{Stressor correlations}: questions about statistical associations among the climatic, biological, environmental, and anthropogenic stressors represented in the ecosystem dataset. Examples include ``What factors correlate with biodiversity?'', ``What are the main correlates of evapotranspiration change across the area?'', and ``What is the relationship between temperature change and biodiversity in the ecosystem?''

\item \textit{Species habitat}: questions concerning habitat suitability, spatial extent, spread, fragmentation, connectivity, model drivers, or reliability for a selected species. Examples include ``Is the kingfisher's habitat fragmented or well connected?'', ``Does temperature influence habitat suitability for kelp?'', and ``How reliable is the habitat assessment for seabass?''

\item \textit{Scenario projection}: questions asking how ecosystem risk could change under one hypothetical modification of one or more stressors. Examples include ``How would ecosystem risk change if tree cover decreased by 50\%?'', ``What would happen if grassland cover increased from 30\% to 50\%?'', and ``What if temperature increased by \(3 \, ^{\circ}\mathrm{C}\) and tree cover decreased significantly?''

\item \textit{Scenario comparison}: questions comparing alternative environmental changes, pressures, or conservation interventions. Examples include ``Which scenario would be worse: a \(2\,^{\circ}\mathrm{C}\) temperature increase or a 20\% precipitation decrease?'', ``Which is more relevant to ecosystem risk: hydrological stress or habitat degradation?'', and ``Which scenario would have the greater effect on ecosystem health: biodiversity decline or grassland loss?''

\item \textit{Data retrieval}: questions regarding the variables available to the DTE or the latest observations for the study area. Examples include ``Which environmental variables are being monitored?'', ``Show me the latest temperature and precipitation values'', and ``Give me the most recent data available for the area.''

\item \textit{Conversation}: greetings, requests for clarification, and other expressions that do not require activation of a specialised module.

\end{enumerate}

These categories are semantically related and may share substantial vocabulary. The terms \textit{temperature}, \textit{biodiversity}, or \textit{tree cover}, for example, can occur in questions about correlations, model importance, scenario simulation, scenario comparison, or data retrieval. A rule-based classifier relying on keywords or mutually exclusive syntactic patterns would consequently require numerous exceptions and could easily route a query according to the variables it mentions rather than the analytical operation requested by the user. 

\begin{algorithm}[ht]
\caption{Query classification and routing}
\label{alg:query_routing}
\begin{algorithmic}[1]
\Require User query \(q\); query-example sets \(\mathcal{E}_1,\ldots,\mathcal{E}_K\) for the \(K\) supported categories; embedding model \(f(\cdot)\)
\Ensure Routed query category \(c^\ast\)

\State Load or compute the cached embeddings \(f(e)\) for all examples \(e \in \mathcal{E}_k\), \(k=1,\ldots,K\)
\State Compute the embedding of the incoming query: \(\mathbf{q}=f(q)\)
\For{each category \(k=1,\ldots,K\)}
    \State Compute cosine similarities between \(\mathbf{q}\) and all embedded examples in \(\mathcal{E}_k\)
    \State Store the highest similarity score for category \(k\):
    \[
    s_k=\max_{e \in \mathcal{E}_k} \cos(\mathbf{q}, f(e))
    \]
\EndFor
\State Identify the most similar category:
\[
c^\ast=\arg\max_k s_k
\]
\If{\(s_{c^\ast}\) is lower than the minimum acceptance threshold}
    \State $c^\ast \leftarrow \text{Conversation}$
    \State Route the query to the conversation module
\Else
    \State Route the query to the analytical module associated with \(c^\ast\)
\EndIf
\State \Return \(c^\ast\)
\end{algorithmic}
\end{algorithm}

The routing mechanism (summarised in Algorithm \ref{alg:query_routing}) focuses on the semantic similarity between the incoming query and representative examples of each task. A large set of example queries is prepared for each category (between 200 and 250 in the Massaciuccoli case). These examples express the same underlying analytical intent through different vocabulary, grammatical structures, levels of detail, and combinations of ecosystem variables. During system initialisation, each example query \(q_{c,i}\), where \(c\) identifies the category and \(i\) the example within that category, is converted into a dense numerical vector using the \texttt{Nomic} text embedder $\mathbf{e}_{c,i}=f(q_{c,i})$, where \(f(\cdot)\) denotes the embedding function. Since the reference queries do not change during ordinary agent operation, their vectors are calculated once and cached. This operation avoids repeatedly embedding the complete reference collection and reduces the latency of subsequent classification requests. 

When a new user query \(q_u\) arrives, it is embedded via \texttt{Nomic} into the same semantic space as the examples, yielding vector $\mathbf{e}_u=f(q_u)$. The router then calculates the cosine similarity between the user-query vector and every cached reference vector:

$$
s(\mathbf{e}_u,\mathbf{e}_{c,i}) =
\frac{\mathbf{e}_u \cdot \mathbf{e}_{c,i}}
{\lVert\mathbf{e}_u\rVert\,\lVert\mathbf{e}_{c,i}\rVert}.
$$

Cosine similarity measures the orientation of the vectors independently of their magnitude. Higher values indicate that the user query and reference example express more similar meanings according to the embedding model. Comparisons with the seven groups of reference queries are performed in parallel because the similarity calculations for one category are independent of those for the others.

For each category, the router identifies the reference example most similar to the user query. The final prediction corresponds to the category containing the globally most similar example:

$$
{c^\ast}
=
\underset{c}{\arg\max}
\left(
\underset{i}{\max}\;
s(\mathbf{e}_u,\mathbf{e}_{c,i})
\right).
$$

The method produces continuous similarity scores and is therefore more flexible than an exclusive set of manually defined rules, even though the final decision assigns the query to one category. Finally, if $\underset{i}{\max}\;
s(\mathbf{e}_u,\mathbf{e}_{c,i})$ falls below a (configurable) minimum similarity threshold (60\% by default) the category is classified as \textit{conversation} and managed through conversational AI (Section \ref{sec:conversation_module}). 

Once the category \({c^\ast}\) has been identified, the query and the classification are passed to the corresponding processing module (next sections). In summary, routing selects among predefined analytical capabilities but does not define a new analytical plan. This query-dependent selection and parameterisation provides flexibility within the supported operations without delegating control over the execution sequence to the LLM.

\subsubsection{Ecosystem drivers module}
\label{sec:ecosystemdrivers}

The ecosystem drivers module identifies the environmental, climatic, biological, and anthropogenic conditions that characterise the areas classified as being at high ecosystem risk. Its purpose is to extract empirical evidence (rather than causal evidence) about which ecosystem features exhibit comparatively elevated values in high-risk cells. The resulting information provides a structured quantitative basis for answering questions such as which variables are most relevant to ecosystem risk or which conditions characterise the most vulnerable areas.

The module operates on the \textit{ecosystem dataset}, augmented with ecosystem risk classification. Risk classification is used as the target variable, while the remaining columns serve as candidate explanatory features. Geographic coordinates are excluded from the analysis.

Algorithm \ref{alg:ecosystem_drivers} summarises the steps of the module. When a query is routed to this module, the in-memory representation of the ecosystem dataset is converted into a tabular structure suitable for statistical processing.  Statistical characterisation is performed independently for each ecosystem feature. Let \(x_j\) denote the values of feature \(j\) across the area. The module first calculates the median of its area-wide distribution, $
m_j = \operatorname{median}(x_j)$. Then it selects the cells classified as \textit{high risk} and calculates the mean value of the same feature within this subset $\overline{x}_{j,H}
=
\frac{1}{n_H}
\sum_{i \in H} x_{ij}
$; where \(H\) denotes the set of high-risk cells and \(n_H\) is their number. 

A feature is retained when its average value in the high-risk cells is equal to or greater than its basin-wide median ($\overline{x}_{j,H} \geq m_j$). 

For every retained feature, the module calculates a relative elevation score 

$$
s_j =
\begin{cases}
\dfrac{\overline{x}_{j,H}-m_j}{m_j},
    & \text{if } m_j>0, \\[6pt]
\overline{x}_{j,H}-m_j,
    & \text{otherwise}.
\end{cases}
$$

This score expresses how strongly the average condition observed in high-risk cells exceeds the central value of the corresponding feature across the study area. The retained features are subsequently ranked in decreasing order of \(s_j\). Features with higher scores represent conditions that are more markedly elevated in high-risk areas, whereas features that do not satisfy the selection criterion are retained separately as variables not distinctly elevated in high-risk areas. In this process, focusing only on the elevated values is consistent with the variables having been pre-transformed to represent stressors, whose \textit{high} values are indeed associated with higher ecosystem risk (Section \ref{sec:harmonised_data}).

\begin{algorithm}[H]
\caption{Ecosystem drivers extraction}
\label{alg:ecosystem_drivers}
\begin{algorithmic}[1]
\Require Ecosystem dataset \(\mathcal{D}\); risk-class column \(Y\); set of stressor variables \(\mathcal{X}=\{x_1,\ldots,x_p\}\)
\Ensure Structured ecosystem-drivers report \(\mathcal{R}\)

\State Exclude geographic coordinates, metadata columns, and the risk-class column from the candidate explanatory variables
\State Select the set of high-risk cells:
\[
H=\{i : Y_i=\textit{high risk}\}
\]
\State Initialise the retained-feature list \(\mathcal{F}^{+}\leftarrow \emptyset\)
\State Initialise the non-distinctive-feature list \(\mathcal{F}^{-}\leftarrow \emptyset\)

\For{each stressor variable \(x_j \in \mathcal{X}\)}
    \State Extract its values across the full basin:
    \[
    x_j=\{x_{1j},x_{2j},\ldots,x_{nj}\}
    \]
    \State Compute the area-wide median:
    \[
    m_j=\operatorname{median}(x_j)
    \]
    \State Compute the mean value in high-risk cells:
    \[
    \overline{x}_{j,H}=\frac{1}{n_H}\sum_{i\in H}x_{ij}
    \]
    \If{\(\overline{x}_{j,H} \geq m_j\)}
        \If{\(m_j>0\)}
            \State Compute the relative elevation score:
            \[
            s_j=\frac{\overline{x}_{j,H}-m_j}{m_j}
            \]
        \Else
            \State Compute the absolute elevation score:
            \[
            s_j=\overline{x}_{j,H}-m_j
            \]
        \EndIf
        \State Add \((x_j,\overline{x}_{j,H},m_j,s_j)\) to \(\mathcal{F}^{+}\)
    \Else
        \State Add \((x_j,\overline{x}_{j,H},m_j)\) to \(\mathcal{F}^{-}\)
    \EndIf
\EndFor

\State Sort \(\mathcal{F}^{+}\) in decreasing order of \(s_j\)
\State Build the structured report \(\mathcal{R}\) containing:
\Statex \hspace{\algorithmicindent} ranked stressors elevated in high-risk cells (i.e., in $\mathcal{F}^{+}$);
\Statex \hspace{\algorithmicindent} their basin-wide median, high-risk mean, and elevation score;
\Statex \hspace{\algorithmicindent} separately report the variables not distinctively elevated under the adopted criterion (i.e., in $\mathcal{F}^{-}$).
\State \Return \(\mathcal{R}\)

\end{algorithmic}
\end{algorithm}

The statistical characterisation of high-risk areas differs from a causal analysis, which would require an appropriate causal design and sufficiently spatio-temporally resolved experimental data, normally unavailable for many ecosystem variables. Instead, this module identifies features that tend to occur together at elevated levels in regions already classified as high risk by a complex model that examined multivariate, non-linear interactions among ecological vulnerabilities and harmful influences. The characterisation serves as the analytical evidence supplied to the answer-building module, which combines it with the original user query and contextual knowledge retrieved from the scientific document collection to finally produce an answer (Section \ref{sec:prompt_builder}). 

\subsubsection{Stressor correlations module}
\label{sec:stressor_correlations}

The stressor correlations module identifies statistical correlations among the stressor variables in the \textit{ecosystem dataset}. It supports questions concerning the relationship between two specific stressors or the variables most strongly associated with a selected ecosystem condition. The resulting associations provide a quantitative background for answering the user's query.

When a query is routed to this module (Algorithm \ref{alg:stressor_correlations}), the LLM is used as an information-extraction component. A prompt is constructed to instruct the LLM to interpret the natural-language request and extract its analytical elements in a structured JSON format (specified in ``Software and data availability''). The variables reported in the JSON document should include the \textit{source variable}, the \textit{target variable}, the requested association \textit{type} or direction (positive/negative), and whether the source or target is a specific stressor or a generic reference to all possible drivers. For example, in the question ``What is the relationship between temperature change and biodiversity?'', temperature change is identified as the source and biodiversity as the target. Conversely, in a query such as ``Which factors are associated with biodiversity?'', the source is interpreted as generic and all eligible ecosystem variables are evaluated against biodiversity.

The extracted variable names are matched to the corresponding columns of the ecosystem dataset. Because users may employ abbreviated, non-technical, or alternative expressions, this normalisation is performed by comparing the embedding of each extracted expression with those of the dataset column names. The column with the highest cosine similarity is selected as the corresponding ecosystem feature. This operation connects natural-language expressions such as ``rainfall'', ``species richness'', or ``tree density'' to the more detailed variable names used in the ecosystem dataset.

The statistical association method is selected according to the data types of the two variables. For two numerical variables $X$ and $Y$, the module calculates the Pearson correlation coefficient $r_{XY}$ over the entire ecosystem dataset. The sign of \(r_{XY}\) indicates the direction of the association, while its absolute value represents its strength. When one variable is categorical, and the other is numerical, the module uses the Pearson correlation ratio \(\eta\), which quantifies how much of the variability in the numerical variable is associated with differences among the categories. When both variables are categorical, their association is measured using Cramér's \(V\), derived from the chi-squared statistic. This automatic selection allows the same module to process different combinations of ecosystem variables without requiring the user to specify the statistical measure. The calculated values are translated into qualitative association classes using commonly used heuristic interpretations (\cite{cohen_1988_statistical}, Table \ref{tab:association_interpretation}).

If the user specifies both a source and a target, the module reports the association between them along with its interpretation. If the query (or the target, likewise) contains a generic source, each eligible ecosystem feature is compared with the selected target, excluding the target itself and the ecosystem-risk classification. 

The resulting associations are organised as source-target pairs containing both the numerical coefficient and its qualitative interpretation. A summary report is eventually supplied to the answer-building module (Section \ref{sec:prompt_builder}).

\begin{algorithm}[H]
\caption{Stressor correlation extraction}
\label{alg:stressor_correlations}
\begin{algorithmic}[1]
\Require User query \(q\); ecosystem dataset \(\mathcal{D}\);
stressor variables \(\mathcal{X}=\{x_1,\ldots,x_p\}\);
embedding model \(f(\cdot)\)
\Ensure Structured stressor-correlation report \(\mathcal{R}\)

\State Use the LLM to identify the source and target variable
expressions, whether either refers generically to all stressors,
and any requested association direction
\State Embed specific variable expressions and match them to
the most similar stressor names using cosine similarity
\State Expand generic expressions to all eligible stressor
variables, excluding coordinates, metadata, and ecosystem-risk
classification
\State Construct the requested source--target pairs,
excluding self-comparisons
\State Initialise the correlation report
\(\mathcal{R}\leftarrow\emptyset\)

\For{each requested source--target pair \((x_s,x_t)\)}
    \State Extract paired valid observations of \(x_s\) and
    \(x_t\) from \(\mathcal{D}\)
    \State Identify the data types of the source and target variables
    \If{both variables are numerical}
        \State Compute the Pearson correlation coefficient:
        \[
        r=\frac{\operatorname{cov}(x_s,x_t)}
        {\sigma_{x_s}\sigma_{x_t}}
        \]
        \State Classify the association according to the sign
        and magnitude of \(r\)
        \State Store the variable pair, \(r\), direction,
        and strength class in \(\mathcal{R}\)
    \ElsIf{one variable is categorical and the other is numerical}
        \State Compute the correlation ratio:
        \[
        \eta=\sqrt{
        \frac{\sum_g n_g(\overline{z}_g-\overline{z})^2}
        {\sum_i(z_i-\overline{z})^2}
        }
        \]
        \Statex \hspace{\algorithmicindent}
        where \(z_i\) denotes the numerical observations,
        \(g\) indexes the categories, and \(n_g\) is the
        number of observations in category \(g\)
        \State Classify the association strength from \(\eta\)
        \State Store the variable pair, \(\eta\), and strength
        class in \(\mathcal{R}\)
    \Else
        \State Compute Cram\'er's \(V\) from the contingency table:
        \[
        V=\sqrt{\frac{\chi^2}{n\min(r_c-1,c_c-1)}}
        \]
        \Statex \hspace{\algorithmicindent}
        where \(\chi^2\) is the chi-squared statistic,
        \(n\) is the paired sample size, and \(r_c,c_c\)
        are the numbers of rows and columns
        \State Classify the association strength from \(V\)
        \State Store the variable pair, \(V\), and strength
        class in \(\mathcal{R}\)
    \EndIf
\EndFor

\State Apply any requested direction filter to Pearson
correlations; identify \(\eta\) and \(V\) as unsigned measures
for which direction filtering is not applicable
\State Add the matched variable names, original user expressions,
and qualitative interpretations to \(\mathcal{R}\)
\State \Return \(\mathcal{R}\)

\end{algorithmic}
\end{algorithm}

\subsubsection{Species habitat module}
\label{sec:species_habitat}

The species habitat module provides the agent interface to the SDM described in Section~\ref{sec:sdm}. It produces an on-demand, model-based characterisation of a focal species within the study area by combining GBIF occurrences with the spatial predictors contained in the \textit{ecosystem dataset}. Through the outputs produced by the SDM, the module supports queries concerning habitat suitability, spatial coverage, fragmentation, connectivity, patch structure, influential environmental predictors, and model reliability. Processing is necessarily conducted on demand because precomputing and caching every possible species distribution would be impractical in biodiversity-rich areas.

When a query is routed to this module (Algorithm \ref{alg:species_habitat}), the LLM acts as an information-extraction component and identifies the species name supplied by the user, which may be either scientific or vernacular. The extracted name is passed to the GBIF taxonomic service to resolve it against accepted species-level taxa. Candidate records are compared with the supplied name using both their scientific and available vernacular names, and the closest accepted match is selected. Its accepted scientific name is then used in the subsequent occurrence query.

The module queries the GBIF occurrence service within the polygon that delimits the study area and retains records that contain geographic coordinates. The study-area polygon and the maximum number of records to retrieve are specified in the agent configuration. Limiting the number of occurrences bounds data transfer and processing times, thereby supporting on-demand answer generation. For the Massaciuccoli case study, exploratory queries returned an average of approximately 100 occurrence records per species; therefore, the maximum was set to 200. The procedure is taxonomically agnostic and can be applied to different biological groups represented in GBIF, provided that the requested name can be resolved to a species-level taxon and that georeferenced occurrences are available within the study area. 

Occurrence data quality control thus comprises taxonomic consistency with the resolved species, availability of geographic coordinates, and spatial consistency with the study-area polygon. These checks constrain the input records, although they cannot prevent individual positional errors, misidentifications, or sampling bias \citep{vanden2015retrieving}.

The spatial predictors required by the SDM are prepared from the ecosystem dataset. Longitude and latitude identify the centres of its harmonised grid cells, while each numerical ecosystem variable is converted into an ESRI ASCII raster layer. The resulting rasters preserve the common extent, resolution, and spatial alignment of the original dataset, and cells without valid information are assigned a designated no-data value. This conversion produces the spatially consistent predictor stack required by the modelling workflow.

The occurrence data and raster predictors are then supplied to the SDM. To this end, the module associates a feature vector with each presence and background location, trains the SDM model, and projects the model onto every cell in the area. The continuous SDM presence probabilities are then converted into a binary presence-absence map by allowing 10\% of the training presences to fall below the suitability cut-off (10th-percentile training-presence logistic threshold), thereby reducing sensitivity to potentially imprecise or environmentally atypical occurrence records \citep{coro2020global}. Cells above or equal to the threshold are assigned 1 and interpreted as suitable habitat, while the remaining cells are assigned 0.

Let \(N\) denote the number of cells in the study area and \(N_s\) the number classified as species presence. Habitat coverage is consequently calculated as $C = 100\frac{N_s}{N}$. Connected-component analysis is then applied to the binary map: adjacent suitable cells are grouped into presence areas (\textit{habitat patches}), and the following measures are extracted: 

\begin{itemize}
\item the total number of connected patches;
\item the percentage of presence cells contained in the largest patch over the total presence cells in the analysed area;
\item the median patch size;
\item the mean minimum border-to-border distance between the largest patch and the remaining patches, divided by the grid-cell size and therefore expressed in cell units.
\end{itemize}

A deterministic rule set converts the spatial statistics into concise descriptions. The rules were extracted by studying the distributions of more than 2,000 marine and terrestrial species across different spatial scales from previous work \citep{coro2016automatic,coro2024climate,coro2025open}. They characterise habitat separation, connectivity, structure, and spread, as reported in Table \ref{tab:habitat_spatial_classification}: a distribution consisting of a single connected patch is considered not fragmented. When several patches are present, low-to-moderate fragmentation with a dominant presence area is deduced if the largest patch contains at least 60\% of the predicted presence cells and the satellite patches are no more than 5 cells (i.e., 5 times the spatial resolution) apart on average. A largest-patch contribution of at least 40\% indicates moderate fragmentation with a dominant area and smaller patches. High fragmentation is deduced when more than 20 patches are detected or when the median patch size is no greater than 2 cells. The remaining configurations are classified as moderately fragmented. 

\begin{algorithm}[H]
\caption{Species habitat information extraction}
\label{alg:species_habitat}
\begin{algorithmic}[1]
\Require User query \(q\); ecosystem dataset \(\mathcal{D}\);
study-area polygon \(P\); GBIF taxonomic and occurrence services;
SDM model \(M_{\mathrm{SDM}}\)
\Ensure Structured species-habitat report \(\mathcal{R}\)

\State Use the LLM as an information-extraction model to identify
the species expression \(u_s\) mentioned in \(q\)
\State Query the GBIF taxonomic service using \(u_s\)
\State Select the accepted species-level taxon whose scientific
or vernacular name most closely matches \(u_s\)
\State Extract the resolved taxon identifier and scientific name
\State Query the GBIF occurrence service using the resolved taxon
and the study-area polygon \(P\)
\State Retain georeferenced occurrence records consistent with
the resolved taxon and the study-area polygon \(P\)
\State Apply the configured maximum number of occurrence records,
if necessary, to keep the computation interactive

\State Extract longitude, latitude, and eligible numerical
ecosystem variables from \(\mathcal{D}\)
\State Convert each eligible ecosystem variable into an aligned
ESRI ASCII raster layer
\State Assign the configured no-data value to cells with
unavailable information
\State Build the raster predictor stack \(\mathcal{G}\),
preserving the spatial extent, resolution, and alignment
of \(\mathcal{D}\)

\State Run the SDM model \(M_{\mathrm{SDM}}\) using the occurrence
records and raster predictors \(\mathcal{G}\)
\State Obtain the continuous habitat-suitability map \(S\),
model-quality diagnostics, and predictor contributions
\State Extract the 10th-percentile training-presence logistic
threshold \(\tau\)
\State Binarise the suitability map over valid study-area cells,
preserving no-data cells:
\[
B_i =
\begin{cases}
1, & S_i \geq \tau,\\
0, & S_i < \tau.
\end{cases}
\]
\State Identify suitable-habitat patches from \(B\)
\State Compute habitat coverage:
\[
C = 100 \frac{N_{\mathrm{suitable}}}{N_{\mathrm{total}}},
\]
\Statex \hspace{\algorithmicindent}
where \(N_{\mathrm{total}}\) is the number of valid
study-area cells
\State Compute the number of suitable patches \(N_p\)

\If{\(N_p=0\)}
    \State Report no suitable habitat predicted
    \State Mark patch-size, dominance, separation, connectivity,
    and structure descriptors as undefined
\Else
    \State Compute \(L\), the percentage of total suitable habitat
    contained in the largest patch
    \State Compute the mean and median patch sizes
    \If{\(N_p>1\)}
        \State Compute \(D\) as the mean minimum border-to-border
        distance from the largest patch to each remaining patch,
        divided by the grid-cell size
    \Else
        \State Mark core-to-satellite separation as not applicable
    \EndIf
    \State Classify habitat fragmentation, separation,
    connectivity, structure, and spread using the rule-based
    thresholds in Table \ref{tab:habitat_spatial_classification}
\EndIf

\State Build the structured report \(\mathcal{R}\) containing:
\Statex \hspace{\algorithmicindent}
resolved species name and taxonomic information;
\Statex \hspace{\algorithmicindent}
number of GBIF occurrences used;
\Statex \hspace{\algorithmicindent}
model diagnostics and predictor contributions;
\Statex \hspace{\algorithmicindent}
habitat suitability and spatial descriptors;
\Statex \hspace{\algorithmicindent}
rule-based habitat classifications and any undefined descriptors.
\State \Return \(\mathcal{R}\)

\end{algorithmic}
\end{algorithm}

Overall, these rules translate complementary spatial properties of the modelled habitat into standardised qualitative descriptions and provide a consistent interpretation layer. Coverage and patch-size statistics characterise the species spread, while patch number, dominance, and separation describe fragmentation, internal structure, and connectivity. Since the thresholds were empirically derived from previous analyses spanning multiple spatial scales, they may be general enough to be applied across different application cases.

After processing, the module collects diagnostics, classifications, interpretations, and the contributions of predictors to the SDM. It assembles the results into a structured report that constitutes the analytical evidence produced by the module. The report is then passed to the answer-building module (Section~\ref{sec:prompt_builder}), which will combine it with the original query and contextual information retrieved from the scientific document collection.

\subsubsection{Scenario projection module}
\label{sec:scenario_projection}

The scenario projection module estimates how the spatial prevalence of high ecosystem risk would change under a hypothetical scenario involving one or more modifications of the ecosystem stressors. It supports questions such as how ecosystem risk could respond to increased temperature, reduced precipitation, declining biodiversity, or combinations of climatic and habitat changes. The module provides a model-based projection relative to the conditions represented in the \textit{ecosystem dataset}.

When a query is routed to this module (Algorithm \ref{alg:scenario_projection}), the LLM is tasked with converting the natural-language scenario into a structured JSON representation (reported in ``Software and data availability'') that includes a list of stressor modifications. For each modification, the JSON document reports the user-specified variable, the original expression from which it was extracted, and the magnitude and unit of the change. A single scenario can therefore involve simultaneous changes in several stressors.

Changes can be expressed as relative variations, such as a 20\% reduction in precipitation, or as absolute variations, such as a \(3\,^{\circ}\mathrm{C}\) increase in temperature. Qualitative expressions without an explicit magnitude, such as ``substantially increased temperature'' or ``reduced biodiversity'', are mapped to a default relative variation of \(+50\%\) or \(-50\%\), respectively. The direction of each change is checked against the linguistic expression to ensure that increases are represented by positive values and decreases by negative values.

The variable names extracted from the query are subsequently matched to the corresponding columns of the ecosystem dataset. As in the stressor correlations module, this normalisation is based on the cosine similarity between the embedding of the user-provided expression and those of the dataset feature names. This matching allows concise expressions such as ``temperature'', ``rainfall'', or ``tree cover'' to be associated with the longer technical names used in the dataset.

The module creates a copy of the ecosystem dataset and applies the requested modifications to the corresponding stressor columns. Let \(x_{ij}\) be the value of stressor \(j\) in spatial cell \(i\). For a relative change \(\delta_j\), expressed as a fraction, the modified value is calculated as

$$
x'_{ij}=x_{ij}+\lvert x_{ij}\rvert\delta_j.
$$

The absolute value ensures that the direction of the requested modification is preserved even when the original stressor value is negative. For an absolute change \(\Delta_j\), the modified value is instead

$$
x'_{ij}=x_{ij}+\Delta_j.
$$

The same transformation is applied to every valid spatial cell, while all stressors not mentioned in the scenario retain their baseline values. The resulting table (\textit{scenario dataset}) represents the ecosystem scenario to be evaluated.

Scenario projection is performed using a Random Forest trained to reproduce the ecosystem-risk classification associated with the baseline \textit{ecosystem dataset}. The stressor variables are used as predictors, and the ecosystem-risk class as the response variable. As a reproducible implementation choice consistent with previous Random Forest applications in ecosystem modelling that optimised predictions of ecosystem data on very similar datasets \citep{masnadi2020relative,Coro2022}, the model uses 200 decision trees by default and a fixed random seed to ensure reproducibility. These parameters are, however, configurable.

\begin{algorithm}[H]
\caption{Scenario projection}
\label{alg:scenario_projection}
\begin{algorithmic}[1]
\Require User query \(q\); ecosystem dataset \(E\); baseline risk labels \(y\);
         feature-name embeddings \(\mathcal{F}\); trained risk model \(M\)
\Ensure Structured report describing the scenario and the projected variation
        in high-risk cells

\State Use the LLM to extract the set of requested stressor modifications
\[
\mathcal{S}=\{(u_j,\delta_j,t_j,e_j)\}_{j=1}^{k},
\]
where \(u_j\) is the stressor expression, \(\delta_j\) its signed variation,
\(t_j\in\{\textsc{relative},\textsc{absolute}\}\) the variation type, and
\(e_j\) the original expression extracted from \(q\)

\ForAll{\((u_j,\delta_j,t_j,e_j)\in\mathcal{S}\)}
    \If{\(\delta_j\) is not explicitly quantified}
        \State Assign the default signed relative variation indicated by the
        qualitative expression
    \EndIf
    \State Check that the sign of \(\delta_j\) is consistent with the stated
    direction of change
    \State Embed \(u_j\) and match it to the ecosystem variable \(v_j\) having
    the highest cosine similarity in \(\mathcal{F}\)
\EndFor

\State Create the scenario dataset \(E' \gets E\)

\ForAll{matched modifications \((v_j,\delta_j,t_j)\)}
    \ForAll{valid spatial cells \(i\)}
        \If{\(t_j=\textsc{relative}\)}
            \State \(x'_{ij}\gets x_{ij}+\lvert x_{ij}\rvert\delta_j\)
        \Else
            \State \(x'_{ij}\gets x_{ij}+\delta_j\)
        \EndIf
    \EndFor
\EndFor

\If{the risk-surrogate model is unavailable or obsolete}
    \State Train a Random Forest \(M\) using the stressors in \(E\) as
    predictors and the baseline risk labels \(y\) as the response
    \State Store \(M\) for reuse in subsequent scenario queries
\EndIf

\State \(\widehat{y}\gets M(E)\) \hspace{1cm} (\texttt{Baseline predictions})
\State \(\widehat{y}'\gets M(E')\) \hspace{0.8cm}
(\texttt{Scenario predictions})

\State \(N_H\gets \sum_i \mathbb{1}(\widehat{y}_i=\text{high})\)
\State \(N'_H\gets \sum_i \mathbb{1}(\widehat{y}'_i=\text{high})\)

\If{\(N_H>0\)}
    \State \(\displaystyle
        \Delta_H\gets100\frac{N'_H-N_H}{N_H}\)
\Else
    \State Mark the relative variation as undefined and report
    \(N'_H-N_H\) as an absolute change
\EndIf

\State Construct report \(R\) containing:
\Statex \hspace{\algorithmicindent}
       the original scenario expressions and their matched stressors;
\Statex \hspace{\algorithmicindent}
       the type, direction, magnitude, and unit of every modification;
\Statex \hspace{\algorithmicindent}
       \(N_H\), \(N'_H\), and \(\Delta_H\);
\Statex \hspace{\algorithmicindent}
       the conditional, model-based interpretation of the projection

\State \Return \((R,N_H,N'_H,\Delta_H)\)
\end{algorithmic}
\end{algorithm}

Once trained, the Random Forest is stored locally and reused for subsequent scenario queries until the ecosystem dataset changes. The trained model is then applied to the hypothetical scenario dataset. Let \(N_H\) denote the number of baseline cells predicted as \textit{high risk}, and \(N'_H\) the number predicted as \textit{high risk} after applying the scenario. The projected relative variation in high-risk cells is calculated as

$$
\Delta_H =
100\frac{N'_H-N_H}{N_H}.
$$

A positive value of \(\Delta_H\) indicates an increase in the number of cells classified as being at high risk, whereas a negative value indicates a decrease relative to the baseline prediction. The reported result, therefore, represents a relative change in the spatial prevalence of the \textit{high risk} class.

The Random Forest serves as a computational surrogate, or statistical emulator, of the previously generated ecosystem-risk classification. It approximates the mapping learned by the complete risk-assessment workflow between ecosystem stressors and resulting risk classes, thereby avoiding the need to rerun the substantially more computationally demanding ecosystem risk model for every user query. The surrogate is not intended to replace the original ecosystem-risk model, identify new risk mechanisms, or provide an independent cell-level risk assessment. Instead, it supports rapid comparisons between a common baseline and hypothetical scenarios by estimating the direction and aggregated magnitude of change in the spatial prevalence of the high-risk class. This intended use requires the surrogate to preserve the overall risk pattern and its aggregate response to stressor modifications rather than reproduce every individual cell label. 

Random Forest surrogates developed for datasets generated using the same ecosystem-risk methodology \citep{vannini2025fair,pavirani2025assessing,bove2026estimating,Pavirani2026} reproduced the original risk classifications with average accuracies above 90\% in training self-tests across different study areas and spatial scales. Moreover, in the proposed DTE, all scenarios are evaluated relative to the predictions of the same trained Random Forest under the baseline conditions. Consequently, systematic approximation errors shared by the baseline and scenario predictions are partly controlled, and the reported quantity represents a relative change in the number of cells assigned to the high-risk class rather than a new absolute estimate of ecosystem risk. 

%The Random Forest serves as a computational surrogate for the previously generated ecosystem-risk classification. It learns the relationship between stressors and the high risk class estimated by the ecosystem-risk model, then projects it onto the scenario dataset. This design enables rapid, on-demand scenario evaluation while retaining the information captured by the more detailed risk-assessment model. The purpose is to estimate reliable, aggregated relative differences in the prevalence of high-risk cells across scenarios, rather than to conduct a complete ecosystem-risk reassessment. 

%The reliability of this approach was evaluated using additional multiscale datasets generated with the same ecosystem-risk assessment model \citep{vannini2025fair,pavirani2025assessing,bove2026estimating,Pavirani2026}.

The module organises the original scenario expressions and the estimated variation in high-risk cells into a structured report, which is eventually provided to the answer-building module to produce an answer (Section~\ref{sec:prompt_builder}). 

\subsubsection{Scenario comparison module}
\label{sec:scenario_comparison}

The scenario comparison module extends the scenario projection procedure described in Section \ref{sec:scenario_projection} to compare two alternative environmental changes, anthropogenic pressures, or conservation scenarios. It supports questions concerning which of two scenarios is associated with a greater projected prevalence of high-risk cells or which intervention produces the more favourable projected outcome. Each scenario may involve modifying one or more stressors.

When a query is routed to this module (Algorithm \ref{alg:scenario_comparison}), the LLM converts the natural-language comparison into a new structured JSON representation (``Software and data availability''). The structure separates the query into scenarios \(A\) and \(B\) and, for each scenario, records the variables to be modified, their original expressions, and the magnitude, direction, and unit of each change. For example, a query comparing a \(2\,^{\circ}\mathrm{C}\) temperature increase with a 20\% precipitation decrease produces one modification for temperature in scenario \(A\) and one for precipitation in scenario \(B\). More complex comparisons can associate several simultaneous changes in stressors with either scenario. The new scenarios correspond to two separate \textit{scenario datasets}, generated through the same process described in the previous section. 

The Random Forest model described in Section \ref{sec:scenario_projection} is subsequently applied to both scenario datasets. Let \(N_H\) denote the number of baseline cells predicted as \textit{high risk}, and \(N_H^{(A)}\) and \(N_H^{(B)}\) the corresponding numbers predicted under scenarios \(A\) and \(B\). The relative variation associated with each scenario is calculated as

$$
\Delta_H^{(k)}
=
100\frac{N_H^{(k)}-N_H}{N_H},
\qquad k\in\{A,B\}.
$$

Both projections are therefore evaluated relative to the same reference condition. Positive values indicate an increase in the spatial prevalence of high-risk cells, while negative values indicate a decrease. Comparing the signs and magnitudes of \(\Delta_H^{(A)}\) and \(\Delta_H^{(B)}\) identifies which scenario produces the largest projected risk increase or, in the case of conservation interventions, the largest projected risk reduction. As in the scenario projection module, the baseline Random Forest serves as a computational surrogate for the ecosystem-risk classification and estimates how the learned risk pattern changes when the stressor values are modified. 

The module organises descriptions of both scenarios and their respective variations in high-risk cells into a structured, paired report. This report is eventually passed to the answer-building module (Section~\ref{sec:prompt_builder}).

\begin{algorithm}[H]
\caption{Scenario comparison}
\label{alg:scenario_comparison}
\begin{algorithmic}[1]
\Require User query \(q\); ecosystem dataset \(E\); baseline risk labels \(y\);
         feature-name embeddings \(\mathcal{F}\); trained risk model \(M\)
\Ensure Structured comparison of the projected ecosystem risks under
        scenarios \(A\) and \(B\)

\State Use the LLM to identify and separate the two alternative scenarios
\State \((q_A,q_B)\gets\Call{ExtractScenarios}{q}\)

\State \((R_A,N_H,N_H^A,\Delta_H^A)
    \gets\Call{ScenarioProjection}{q_A,E,y,\mathcal{F},M}\) 
\State \((R_B,N_H,N_H^B,\Delta_H^B)
    \gets\Call{ScenarioProjection}{q_B,E,y,\mathcal{F},M}\)

\If{\(\Delta_H^A\) and \(\Delta_H^B\) are defined}
    \State \(D_{A,B}\gets\Delta_H^A-\Delta_H^B\)

    \If{\(D_{A,B}>0\)}
        \State \(c\gets\) ``scenario \(A\) produces the larger increase
        in high-risk cells''
    \ElsIf{\(D_{A,B}<0\)}
        \State \(c\gets\) ``scenario \(B\) produces the larger increase
        in high-risk cells''
    \Else
        \State \(c\gets\) ``the scenarios produce equivalent projected
        changes in high-risk cells''
    \EndIf
\Else
    \State \(D_{A,B}\gets N_H^A-N_H^B\)
    \If{\(D_{A,B}>0\)}
        \State \(c\gets\) ``scenario \(A\) produces more high-risk cells''
    \ElsIf{\(D_{A,B}<0\)}
        \State \(c\gets\) ``scenario \(B\) produces more high-risk cells''
    \Else
        \State \(c\gets\) ``the scenarios produce the same number of
        high-risk cells''
    \EndIf
\EndIf

\State Construct report \(R\) containing:
\Statex \hspace{\algorithmicindent}
       the descriptions and matched stressor modifications in \(R_A\) and \(R_B\);
\Statex \hspace{\algorithmicindent}
       the common baseline number of high-risk cells \(N_H\);
\Statex \hspace{\algorithmicindent}
       the projected values \(N_H^A\) and \(N_H^B\);
\Statex \hspace{\algorithmicindent}
       the projected variations \(\Delta_H^A\) and \(\Delta_H^B\);
\Statex \hspace{\algorithmicindent}
       the difference \(D_{A,B}\) and comparison statement \(c\)

\State \Return \(R\)
\end{algorithmic}
\end{algorithm}

\subsubsection{Data retrieval module}
\label{sec:data_retrieval}

The data retrieval module answers questions concerning the variables available to the DTE and their latest aggregated values for the study area. Example requests include asking for the latest environmental observations, the variables representing a particular ecosystem component, or a complete summary of the currently available data (``Software and data availability''). 

Once the query has been classified as a data request, the module retrieves the \textit{live} and \textit{ecosystem} datasets (Algorithm \ref{alg:data_and_conversation}). A matching process retrieves variables and descriptions that match the user's request. The variable names and their averaged values across the study area are converted into a structured textual representation, with one variable-value pair per line. The names are preserved as reported in the source table so that units, temporal references, and aggregation semantics remain explicit.  If the requested information is unavailable, the module reports that it is not represented in the connected data sources. The original query is subsequently passed to the answer-building module to select and present the relevant variables from the retrieved information (Section~\ref{sec:prompt_builder}). 

\begin{algorithm}[H]
\caption{Data retrieval and conversation handling}
\label{alg:data_and_conversation}
\begin{algorithmic}[1]
\Require User query \(q\); routed category \(c\); ecosystem dataset \(E\);
         live dataset \(L\); document collection \(\mathcal{D}\)
\Ensure Evidence package \(R\) for the prompt-building module

\If{\(c=\textsc{data retrieval}\)}
    \State Use the LLM to convert \(q\) into a structured request
    \[
    J\gets\Call{ExtractDataRequest}{q},
    \]
    specifying the requested variables, information source, temporal
    reference, and whether variable descriptions or current values are required

    \If{\(J\) requests the variables available to the DTE}
        \State Extract variable names and descriptions from the metadata of
        \(E\) and \(L\)
        \State \(V\gets\Call{ListAvailableVariables}{E,L}\)
    \Else
        \ForAll{variable expressions \(u_j\) extracted into \(J\)}
            \State Match \(u_j\) to the most similar variable name or
            description in \(E\) and \(L\) (using a 60\% similarity threshold)
            \If{the matched variable is available in \(L\)}
                \State Retrieve its latest processed value, unit, temporal
                reference, and variation from the previous period
            \ElsIf{the matched variable is available in \(E\)}
                \State Retrieve its description and the relevant harmonised
                ecosystem information
            \Else
                \State Mark the requested variable as unavailable
            \EndIf
        \EndFor
    \EndIf

    \State Construct \(R\) containing the requested variable names,
    descriptions, values, units, temporal references, data sources, and
    explicit indications of unavailable information

\ElsIf{\(c=\textsc{conversation}\)}
    \State Preserve the original user query without extracting analytical
    variables or parameters

    \State Retrieve document passages relevant to \(q\), when contextual
    scientific information is required:
    \[
    K\gets\Call{RetrieveContext}{q,\mathcal{D}}
    \]

    \State Construct \(R\) containing \(q\), the retrieved context \(K\), and
    an instruction to produce a general conversational response 

    \State Specify that claims in the answer about the DTE or the study ecosystem must be
    supported by the retrieved context and that unavailable numerical
    evidence must not be inferred
\EndIf

\State \Return \(R\)
\end{algorithmic}
\end{algorithm}

\subsubsection{Conversation module}
\label{sec:conversation_module}

The conversation module handles greetings, clarification requests, general questions, and other expressions that do not require a specialised scientific module (Algorithm \ref{alg:data_and_conversation}). It provides a fallback mechanism for requests outside the analytical operations explicitly implemented by the DTE, thereby preserving conversational continuity and exposing the broader agentic capabilities of the underlying LLM.

A query is assigned to this module either when similarity-based classification identifies it as conversational or when none of the specialised categories exceeds the minimum score for the unknown-query category (Section~\ref{sec:query_routing}). Therefore, unrecognised requests are forwarded to the LLM, which can use its background, pre-trained knowledge and reasoning skills to respond to greetings, provide definitions or clarifications, and address general questions.

Unlike the specialised modules, the user's query is passed directly to the common retrieval and answer-generation workflow implemented by the answer builder, which combines it with the documentary context to provide a scientifically grounded conversational answer (next section).

\subsection{Answer builder}
\label{sec:prompt_builder}
The answer builder module (Figure \ref{fig:reactive_agent}, right-hand side) constitutes the interface between the analytical modules and the LLM. Its purpose is to preserve a clear separation between the production of scientific evidence and its communication: numerical values, statistical associations, model diagnostics, and scenario projections are produced by the specialised modules, whereas the LLM is responsible for interpreting this evidence and presenting it in a scientifically cautious, query-specific form. 

After the routing and execution of a specialised module, the answer builder collects three main inputs (Algorithm \ref{alg:answer_builder}):

\begin{enumerate}
\item the original user query;
\item the structured analytical evidence produced by the selected module; and
\item contextual information retrieved from the scientific-document collection.
\end{enumerate}

The analytical evidence depends on the query category and the related module (see previous sections). Therefore, it may include ranked ecosystem features, correlation coefficients and their qualitative interpretation, SDM diagnostics and habitat descriptors, projected variations in high-risk cells, comparisons between alternative scenarios, or values retrieved from the knowledge sources. This evidence is treated as the authoritative source for quantitative statements in the answer.

The original query is used to retrieve the most relevant document chunks from the RAG knowledge base described in Section~\ref{sec:documents}. These documents may provide complementary ecological and methodological context that can support the interpretation of the analytical results. However, the LLM will be instructed not to replace, modify, or contradict the numerical evidence produced by the selected module through contextual information. If the analytical output does not contain the requested information, the LLM is instructed not to infer or fabricate variables, values, rankings, spatial patterns, species-specific results, ecological mechanisms, or management recommendations. The LLM will state knowledge limitations rather than complete the answer speculatively. 

Generally, the LLM is instructed to communicate uncertainty in results through concise, evidence-sensitive advice rather than by systematically exposing all underlying statistical diagnostics. The answer builder uses the information supplied by each analytical module to qualify weak associations, limited occurrence support, uncertain model reliability, and scenario projections that are conditional on the available data and model assumptions. The LLM is instructed to avoid causal or definitive language when the evidence is correlational, incomplete, or model-derived, and to report relevant limitations when they materially affect the interpretation. This strategy is intended to preserve accessibility for non-specialist stakeholders, for whom a complete presentation of uncertainty measures could make answers unnecessarily complex. More detailed diagnostics can nevertheless be provided upon explicit request by the user.

The answer builder module orchestrates three main logical components, explained below: agent behaviour, prompt building, and answer building. 

\paragraph{Agent-behaviour component:} Defines the identity, expertise, communication style, and epistemic constraints applied across the category-specific templates. Agent behaviour is defined within a configuration file. It is logically independent of the analytical modules and can be replaced without modifying the modules. By default, the DTE adopts the audience-agnostic behaviour of an ``expert ecosystem scientist, ecological risk analyst, and helpful conversational assistant''. It communicates model results using scientifically cautious language, distinguishes associations from causal relationships, and explicitly identifies model-based results when appropriate. 

Agent behaviour can be adjusted to communicate differently depending on the intended users. For example, a configuration for monitoring authorities could emphasise concise indicators, changes from reference conditions, and decision-relevant implications; a scientific configuration could provide greater methodological detail, uncertainty, and model limitations; and a citizen-oriented configuration could use plain language, minimise technical terminology, and explain ecological concepts before presenting the results.

\paragraph{Prompt building component: }
For a query assigned to category \(c\), the assembled prompt can be represented conceptually as

$$
P_c =
T_c\!\left(
Q,\,
E_c,\,
C_Q,\,
B,\,
R_c
\right),
$$

where \(Q\) is the original query, \(E_c\) is the evidence produced by the corresponding analytical module, \(C_Q\) is the contextual information retrieved for the query from the RAG knowledge base, \(B\) defines the agent's behaviour, and \(R_c\) contains the category-specific interpretation and response rules. The function \(T_c\) represents the prompt template associated with category \(c\). The prompt-building component constructs a structured virtual prompt object that keeps these elements separate. 

The category-specific interpretation and response rules \(R_c\) are summarised in Table \ref{tab:prompt_strategies}. These concern the management of correlation directions, the sign of scenario variations, the distinction between predictor contributions and causal drivers, and the interpretation of model-quality indicators. They are necessary because each query category requires a different relationship between analytical evidence, contextual information, and response instructions. The DTE stores them as external text files that also contain query-and-answer examples, allowing the instructions to be inspected and revised independently. 

\paragraph{Answer building component:} The prompt object from the previous component is passed to the answer building component, which finally assembles the prompt, sends it to the LLM and forwards the answer to the user. The assembled prompt $P_c$ instructs the LLM to (i) answer the original query, (ii) select only the relevant parts of the analytical report, (iii) preserve numerical values and their units, and (iv) avoid exposing internal JSON structures, template fields, prompt instructions, or processing steps. Moreover, it includes the category-specific rules \(R_c\) to correctly manage the data provided by the analytical module used. The prompt also instructs the LLM to use the contextual documents to possibly explain why an observed association or model result may be ecologically meaningful, and specifies that they should not be used to alter the calculated result. It also includes the query-and-answer examples from the prompt object to demonstrate the expected structure and level of detail, as well as general instructions for managing uncertainty in results. 

Finally, $P_c$ is submitted to the LLM, which returns a natural-language response as the user-facing output. The final response constitutes a query-specific verbalisation of the available evidence: its quantitative content derives from the analytical modules, its ecological interpretation is supported by the document collection, and its tone and level of detail are controlled by the selected agent behaviour.

\begin{algorithm}[H]
\caption{Answer building}
\label{alg:answer_builder}
\begin{algorithmic}[1]
\Require User query \(q\); query category \(c\); analytical evidence \(R\);
         document collection \(\mathcal{D}\); behavioural profile \(B\);
         category-specific prompt templates \(\mathcal{T}\);
         conversation history \(H\); language model \(M_{\mathrm{LLM}}\)
\Ensure Final natural-language answer \(a\)

\State Retrieve the document passages most relevant to the user query:
\[
K\gets\Call{RetrieveContext}{q,\mathcal{D}}
\]

\State Select the answer-building template associated with the routed category:
\[
T_c\gets\Call{SelectTemplate}{c,\mathcal{T}}
\]

\State Load the behavioural instructions \(B\), defining the agent as an
ecosystem scientist, ecological-risk analyst, and helpful conversational
assistant

\State Initialise prompt \(P\) with \(B\) and \(T_c\)

\State Add the original user query \(q\) and the relevant conversation
context extracted from \(H\) to \(P\)

\If{\(c\neq\textsc{conversation}\)}
    \State Add the structured analytical evidence \(R\) to \(P\)
    \State Instruct the LLM to preserve the reported variable names,
    numerical values, units, directions, and uncertainty information
\Else
    \State Add the conversational evidence package \(R\), when available
\EndIf

\If{\(K\neq\emptyset\)}
    \State Add \(K\) as complementary scientific and ecosystem-specific
    context
    \State Instruct the LLM to use \(K\) to interpret, rather than replace,
    the analytical evidence
\EndIf

\State Add category-specific interpretation constraints to \(P\)

\If{\(c=\textsc{ecosystem drivers}\)}
    \State Instruct the LLM to describe empirical associations without
    inferring causality
\ElsIf{\(c=\textsc{stressor correlations}\)}
    \State Instruct the LLM to distinguish statistical association from
    causal influence
\ElsIf{\(c=\textsc{species habitat}\)}
    \State Instruct the LLM to report only the requested habitat properties
    and qualify conclusions using the model diagnostics
\ElsIf{\(c=\textsc{scenario projection}\)}
    \State Instruct the LLM to present the result as a conditional,
    model-based projection relative to the baseline
\ElsIf{\(c=\textsc{scenario comparison}\)}
    \State Instruct the LLM to compare both scenarios against their common
    baseline without introducing unsupported mechanisms
\ElsIf{\(c=\textsc{data retrieval}\)}
    \State Instruct the LLM to preserve data provenance, units, and temporal
    references and to report unavailable information explicitly
\ElsIf{\(c=\textsc{conversation}\)}
    \State Instruct the LLM to answer conversationally and avoid unsupported
    DTE-specific numerical claims
\EndIf

\State Add general answer constraints requiring scientific caution,
relevance to \(q\), terminological consistency, and explicit acknowledgement
of missing or uncertain information

\State Generate the candidate answer:
\[
a\gets M_{\mathrm{LLM}}(P)
\]

\State Return \(a\) to the user through the active interface

\State \Return \(a\)
\end{algorithmic}
\end{algorithm}

\subsection{Agent configuration and user interfaces}
\label{sec:user_interfaces}

The proposed software is available as a software suite for rapidly constructing conversational DTEs with enhanced cross-platform portability (``Software and data availability''). The suite is independent of the application case. It separates the reusable analytical and conversational architecture from the data, geographic boundaries, documents, and behavioural instructions associated with a particular ecosystem.

A new application can be configured when four principal resources are available:

\begin{enumerate}
\item a harmonised, spatially explicit dataset representing the ecosystem stressor variables (the \textit{ecosystem dataset}, Section \ref{sec:harmonised_data});
\item a live-data source, with related connectors, that can be updated at an application-dependent frequency (the \textit{live dataset}, Section \ref{sec:live_data});
\item a collection of scientific papers and other authoritative documents describing the ecosystem, its ecological and anthropogenic components, and the pressures acting upon it (the background knowledge document collection, Section \ref{sec:documents}); and
\item a polygonal representation of the study area in Well-Known Text (WKT) format.
\end{enumerate}

The update frequency of the live data is not fixed by the framework and can therefore range from near-real-time sensor acquisition to daily, monthly, or annual aggregation, depending on the ecosystem and application requirements. A configuration file allows the developer to set the module-dependent parameters and specify the locations of the data listed above.

Application behaviour is controlled through external text files. These include the example queries used for semantic routing, the information-extraction templates, and the answer-generation prompts. The information-extraction templates define general rules that can be reused across study areas but should be evaluated for effectiveness in new applications. Ecosystem-specific terminology, background information, variables, response styles, and stakeholder requirements can be introduced in the answer-generation prompts without modifying the analytical code. This text-based configuration facilitates rapid adaptation, transparent inspection of the agent instructions, and iterative behavioural refinement.

Users can run the framework directly from the software suite and interact with the DTE through its conversational command-line chat interface. The chat allows users to interact with a stateless version of the DTE that uses the ecological and analytical context pertinent to the current user's query. This chat can be embedded in a Web application, messaging platform, dashboard, or other service-oriented interface. Direct execution is useful for testing and scientific experimentation, while service integration would make the same analytical functions accessible to non-technical users without exposing the underlying software environment.

The LLM backend is independently configurable. The DTE can consequently use different LLMs, provided that they are exposed through a compatible inference service, without changing the analytical modules. This separation allows deployment managers to balance response quality, inference time, memory requirements, energy consumption, and hardware availability. The current implementation supports all Ollama service models \citep{marcondes2025using} and the Google Gemini Enterprise Agent Platform \citep{google2026geminienterprise} via the bridge described in Section \ref{sec:compared_configurations}, by changing only the model name and the service IP address in the configuration file. 

The framework adaptation to the Massaciuccoli data was named \textit{Massaciuccoli DTE}. This customisation connected the general software architecture to the Massaciuccoli ecosystem dataset, live data flows, study-area polygon, scientific document collection, and case-specific prompt templates. Its behavioural configuration was intended to support questions from scientists, environmental stakeholders, public administrations, and citizens while retaining the audience-neutral behaviour of an expert ecosystem scientist, ecological risk analyst, and helpful conversational assistant. For the scope of the present paper, it was tested on the CNR service infrastructure described in Section \ref{sec:live_data}. This infrastructure provided the computational and distributed storage services that connected the agent to processed observations and external data sources. The testing environment used an NVIDIA RTX~3080 GPU (10 GB VRAM). Across the tests performed, answers were generated in less than 10~s on average, with response times ranging from approximately 1 to 20~s, depending on the requested operation. 

While the current software suite is a constantly updated and enhanced project, the Massaciuccoli DTE version used for the present evaluation was released as a ready-to-use Docker container snapshot to ensure the repeatability and reproducibility of the results (``Software and data availability''). Containerisation packaged the application, runtime libraries, prompt templates, and software dependencies into a versioned execution environment that directly started the textual chat interface. The default configuration of this snapshot uses a CPU-based Llama 3.2 3B LLM to minimise computational requirements while maintaining response times averaging below 2 minutes on a modern laptop. 

The Massaciuccoli DTE was also connected to a public Telegram bot (\texttt{@massaciuccoli\_bot}, \cite{coro2026massaciuccolibot}). Telegram provided a familiar interface that did not require users to install the DTE software or understand its underlying data and modelling architecture. Citizens and stakeholders could submit natural-language questions from mobile or desktop devices and receive the same evidence-supported responses generated by the analytical and prompt-building workflow. The bot served as a free, publicly accessible service for communities and municipalities surrounding the Massaciuccoli Lake basin. The chatbot was presented during a public event \citep{cnrnews2025massaciuccoli}, which brought together research organisations, national and local institutions, environmental organisations, and local administrations to discuss ecosystem restoration and the future management of the lake. Participants tested the bot and discussed its potential to explore risk areas, interpret possible environmental changes, and support conservation and restoration strategies. The bot is currently active during public events involving local citizens, authorities, and scientists, facilitating dialogue among these groups to agree on shared actions or clarify ongoing regulations. These public demonstrations supported stakeholder engagement and exploration of the system’s capabilities. However, they were not meant as a formal validation of its robustness or of the suitability of its answers for management decisions.

These complementary options to access the DTE address different levels of use. The open software suite supports reproducibility and the quick development of new DTEs; the customised repository provides an ecosystem-specific scientific application; the Docker container enables portable, reliable operational deployment; and the Telegram bot lowers the technical barrier to public and stakeholder access.

\subsection{Evaluation methodology}
\label{sec:evaluation_methodology}

The Massaciuccoli DTE was used as the evaluation testbed. The \textit{reactive} DTE architecture was compared with an \textit{agentic} AI baseline using a benchmark of 110 questions representative of the operations supported by the system. These included (i) identifying ecosystem drivers, (ii) analysing correlations among stressors, (iii) characterising species habitats, (iv) projecting and comparing ecosystem-risk scenarios, (v) retrieving data, and (vi) answering general ecosystem-related questions. 

The same target behaviour was assigned to all configurations: providing concise, precise answers from an ecosystem scientist's perspective. For the 110 benchmark questions, intermediate query interpretations and analytical calculations were checked for consistency with the intended requests and the implemented models. These checks confirmed that the extracted parameters reflected the requested analyses and that the calculations adhered to the specified procedures. This verification concerned the correct interpretation and execution of the evaluated cases but did not include an independent validation of ecological predictions or a separate benchmark of extraction accuracy. The experiment thus assessed the final-answer quality of complete configurations.

\subsubsection{Agentic-AI baseline}

\begin{figure}
    \centering
    \includegraphics[width=1\linewidth]{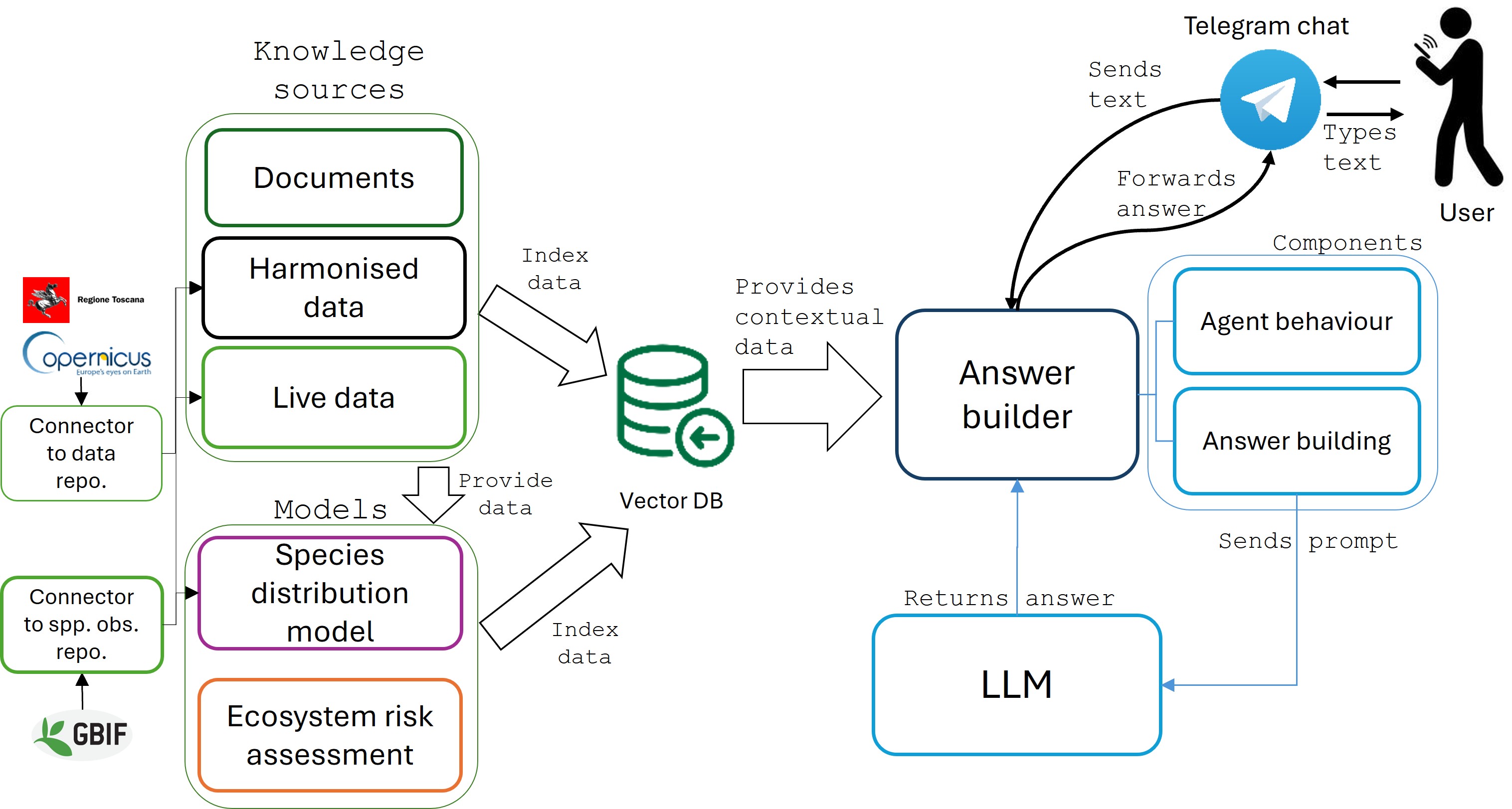}
    \caption{Agentic baseline architecture, using the study-specific definition of LLM-led interpretation of retrieved ecosystem information and previously generated model outputs.}
    \label{fig:agentic_agent}
\end{figure}

The baseline DTE used an agentic architecture in the broader, study-specific sense defined in the Introduction (Figure \ref{fig:agentic_agent}): the LLM was responsible for interpreting ecosystem evidence, identifying relevant relationships, approximating simulated scenarios, and constructing the response. Its evidence sources comprised the twelve-document scientific collection, the ecosystem dataset containing georeferenced stressor values and previously calculated risk classifications, and a remotely accessible table of updated observations. The baseline accessed model-derived information rather than executable versions of the scientific models. 

It was essentially a version of the reactive agent deprived of the analytical modules. In fact, unlike the reactive architecture, the agentic baseline did not execute the query-specific analytical workflows described in Section \ref{sec:modules}. Instead, its answers relied on the LLM's interpretation of the evidence supplied through retrieval and contextual prompts. The comparison thus assessed the complete configurations, including their distinct mechanisms for producing and interpreting analytical evidence, rather than isolating orchestration autonomy under identical access to tools. 

Operationally, the baseline was an LLM-led retrieval-augmented question-answering system, also asked to perform simulations and data analyses on the entire ecosystem dataset via the LLM when needed. Notably, data analyses in this case were conducted through the LLM internal prediction mechanisms rather than through structured analytical modules. The greater flexibility of this architecture compared to the reactive architecture concerned a simpler implementation and the interpretation and narrative synthesis of the supplied evidence, whereas retrieval and access to updated observations were controlled by application logic. The term \textit{agentic} thus follows the broader, study-specific usage introduced above.

The implementation used Python, a Flask conversational interface, LangChain, Ollama, and a persistent Chroma vector database (``Software and data availability''). Scientific documents were divided into non-overlapping chunks of up to 1,200 characters and embedded using \texttt{Nomic}. As in the reactive configuration, retrieval selected the six nearest chunks and retained candidates with squared Euclidean distance no greater than 0.6. Duplicate passages were removed, and the retained chunks were ordered by increasing distance. Instead, the ecosystem dataset records were provided entirely to ensure that analyses could be conducted on all records. To ensure the dataset did not exceed the LLM's maximum token context length, it was subsampled to 1500 rows while maintaining the proportions of low-, high-, and medium-risk cells. This processing resulted in a dataset that enabled the agent to infer correlations and projections consistent with those of the reactive architectures. The LLM received these passages together with the current query and behavioural instructions. Requests were processed independently, without conversation history. Additionally, the agent compared the query embedding with reference expressions for requests involving live data. For these requests, a cosine similarity above 0.6 triggered an HTTP request to the live data, and the agent included these observations in the response by appending all table rows to the answer's context before generating the response. This operation was controlled by application logic.

In summary, the comparison between the proposed reactive agent and the agentic-AI baseline assessed two alternative allocations of reasoning: predominantly within the LLM for the agentic baseline, and predominantly within specialised external modules for the reactive architecture.

\subsubsection{Compared configurations}
\label{sec:compared_configurations}
Since the quality of the answers depended on the LLM's capabilities, a sensitivity analysis was conducted by comparing answers across two agentic and four reactive \textit{configurations} after varying the LLM used. 

The agentic architecture was tested when using the Llama~3.2~3B and Gemini~3.5 Flash LLMs. These two configurations used the same application pipeline, evidence sources, retrieval procedure, and prompt-construction logic. Gemini was accessed through an Ollama-compatible bridge that translated chat and generation requests into Google Vertex AI Gemini API requests and converted the responses into Ollama-compatible output (``Software and data availability''). This allowed the application to change the language model backend via configuration without altering its retrieval or evidence preparation workflow. The bridge did not configure Google Search grounding, URL context retrieval, or other web search tools.

The reactive architecture was tested with the Llama~3.2~3B, Llama~4 Scout, Qwen3~8B, and Gemini~3.5 Flash LLMs. All configurations received the same benchmark questions and were instructed to adopt the same concise scientific-expert behaviour. Including Gemini and Llama 3.2 in both architectures enabled comparisons between configurations using the same underlying language model to manage analytical operations, evidence preparation, and architecture-specific prompts. Moreover, testing diverse LLMs from different providers, with increasing numbers of parameters within the reactive architecture, enabled assessment of how much answer quality varied across models. In the reactive architecture, this comparison was particularly relevant because the analytical logic was implemented independently of the LLM; consequently, the choice of language model is expected to influence performance less strongly than in the agentic-AI baseline.

LLM generation and retrieval settings were tuned using representative questions from each task to balance response completeness and latency for conversational deployment. The tuning operation consisted first of selecting the questions that retrieved the maximum configured number of relevant documentary chunks. Then, context and generation budgets were adjusted to accommodate the evidence, avoid visibly truncated responses, and retain acceptable response times. Eventually, the local context window setting was 4,096 tokens - plus the ecosystem data token count in the case of the agentic architectures -, the temperature was 0, and retrieval returned at most 6 chunks that satisfied the distance cutoff. Maximum generation lengths were instead LLM-specific: 512 tokens for Llama~3.2~3B, 1,024 for Qwen3~8B, and 1,536 for both Llama~4 Scout and Gemini~3.5 Flash. Requests used the Ollama-compatible \texttt{/api/generate} interface; the Gemini bridge mapped the generation limit to the provider’s output-token setting and programmatically enforced the local context setting. These deployment-oriented choices supported comparison of complete configurations.

\subsubsection{Evaluation dimensions}
Each answer from the Massaciuccoli DTE was assessed across seven complementary dimensions:

\begin{enumerate}
\item \textit{Correctness}: the extent to which the answer accurately addressed the question and was consistent with the available evidence;
\item \textit{Scientific narrative}: the ability to organise the answer as a coherent scientific explanation;
\item \textit{Quantitative grounding}: fidelity to the numerical, statistical, and modelling information available to the DTE;
\item \textit{Ecological storytelling}: the ability to explain the ecological processes and interactions underlying the reported result;
\item \textit{Hallucination control}: the avoidance of unsupported, unreliable, or invented claims;
\item \textit{Local ecological detail}: the appropriate use of knowledge specific to the Massaciuccoli Lake basin; and
\item \textit{Fluency}: the linguistic quality, clarity, and naturalness of the response.
\end{enumerate}

These dimensions were selected to distinguish factual and quantitative reliability from explanatory richness. For example, an answer could provide a fluent and ecologically plausible narrative while remaining weakly grounded in the numerical results. Conversely, an answer could reproduce the analytical evidence accurately but provide limited ecological interpretation. 

The distinction between statistical associations and causal relationships was assessed within correctness, quantitative grounding, and hallucination control by both evaluators. In particular, they checked whether answers presented correlations, predictor contributions, or conditional scenario projections as established causal effects. Causal interpretation was thus an embedded assessment criterion, although it was not scored as a separate dimension.

It is worth noting that query classification accuracy was not included among the evaluation assets because, on a testbed of 2000 queries automatically generated by ChatGPT 5.5 based on the query class descriptions, classification accuracy was 100\%. Moreover, real-user queries to evaluate query classification and parameter extraction within the selected modules were unavailable. They will be included in future work with a specific tuning study for query routing, which will necessarily require a dedicated, substantial funding effort to involve experts and annotators. 

The answers were assessed by two evaluators. The \textit{human} evaluator was a non-author domain expert with several years of field experience in the Massaciuccoli Lake basin and detailed knowledge of its ecology, conservation pressures, and management context. The expert had previously collaborated with the authors in public science communication concerning the basin. Recruiting additional evaluators with comparable site-specific expertise was impractical within the study. The expert was able to evaluate both the scientific plausibility of the responses and their consistency with local ecological conditions, knowing that the answers had been generated by automated systems but without knowing architectural details. This evaluation supported the study’s exploratory objective of obtaining a site-specific expert appraisal of the implemented configurations. The expert scored the generated responses using the specified evaluation dimensions.

To complement this necessarily limited human assessment with a systematic and independent evaluation, a second evaluation was performed using ChatGPT 5.5 at the medium reasoning level (\textit{automated} evaluator). The model was instructed to use the same dimensions and scoring scale as the human expert, without access to the expert's assessment. The medium reasoning level was selected because it balanced comparative judgement, careful examination of the answers, and evaluation cost \citep{openai_o3mini_2025,openai_gpt55_2026}. A structured table containing all questions and answers across the agents was uploaded to ChatGPT, and a prompt (reported in ``Software and data availability'') was used to task it with evaluating each question-answer pair for each agent separately. This operation was conducted both via the API and the web interface to ensure the results were consistent.

The quality of each answer across the dimensions was assessed using a five-level Likert scale, converted to percentages as follows: \textit{very low}, 0\%; \textit{low}, 25\%; \textit{moderate}, 50\%; \textit{high}, 75\%; and \textit{very high}, 100\%. Moreover, for each configuration and evaluator, the score for a dimension was calculated as the average across the related question set. Finally, an overall score for each DTE configuration was obtained by averaging the individual answer scores across all seven dimensions. 

The evaluation analysed both answer-level assessments and the overall score to identify the strengths and weaknesses of each configuration, based on the views of the automated and expert evaluators. Two-sided Wilcoxon signed-rank tests were used at a significance level of \(\alpha=0.05\) to examine systematic differences in the paired answer-level scores. When comparing two DTE configurations, the scores were paired by benchmark question: for each of the 110 questions, the answer-level overall score from one configuration was compared with that from the other. These pairwise tests were performed separately for the human and automated evaluators. Consequently, the tests assessed whether one configuration tended to receive systematically higher or lower answer-level scores than another across the same set of questions.

Differences between the human and automated evaluators were assessed in a similar manner. For each DTE configuration, the two evaluators’ scores were paired for each evaluated answer, and a two-sided Wilcoxon signed-rank test was applied to determine whether one evaluator systematically tended to assign higher or lower answer-level scores than the other. An additional overall comparison was conducted across the paired assessments from all configurations. The Wilcoxon tests were selected because the scores originated from an ordinal scale and no assumption of normally distributed score differences was required. A non-significant result was interpreted conservatively as insufficient evidence of a systematic difference in paired scores, rather than evidence of evaluator equivalence or agreement on individual answers.
%A non-significant result was interpreted conservatively as evidence of no systematic differences in scoring.

\section{Results}
\label{sec:results}
This section presents the results of the evaluation methodology. First, a general comparison of the outputs of the evaluated agent configurations is presented (Section \ref{sec:qualitativecomparison}). Then, a numerical performance assessment across the evaluation dimensions is reported (Section \ref{sec:performance_results}). Third, the differences in the agents' overall scores are analysed (Section \ref{sec:configuration_differences}). Finally, inter-evaluator similarities and differences between evaluators are measured (Section \ref{sec:evaluator_agreement}). 

\subsection{General comparison}
\label{sec:qualitativecomparison}

The evaluated agents displayed markedly different answer-building behaviours. Representative examples are reported in Table \ref{tab:examples} and as a Microsoft Excel workbook in ``Software and data availability'', along with another workbook reporting all queries and answers. The agentic configurations generally produced extensive scientific narratives, autonomously connecting the query to ecological mechanisms and management implications. However, these explanations often exceeded the quantitative information available to the DTE, consequently generating hallucinations. Conversely, the reactive configurations produced more concise answers that were more closely constrained by the analytical modules' outputs, with the overall effect of reducing hallucinated content.

These differences were evident in questions about the ecosystem-risk drivers. For example, when asked ``how does biodiversity influence ecosystem risk?'', the agentic Gemini configuration described biodiversity as the ecosystem's ``structural buffer and resilience engine'' and discussed functional redundancy and ecological tipping points. The agentic Llama~3.2 configuration stated that species richness was positively and significantly related to risk, even reporting a \(p<0.001\) result that was not supported by the available analysis. The reactive agents instead reported biodiversity as one of several conditions characterising high-risk cells and placed it within the ranking produced by the ecosystem-drivers module. 

A similar contrast emerged for correlations among stressors. In response to the question ``how does tree cover affect species richness'', the agentic Gemini configuration described tree cover as a fundamental driver and proposed several plausible ecological mechanisms, despite the absence of a clear direct correlation in the data. The reactive Gemini agent explicitly reported the neutral direct relationship and explained that this result did not imply ecological irrelevance, since tree cover could still interact with habitat structure, productivity, microclimate, and land use. Thus, the reactive answer preserved the statistical result while using ecological context to explain its limitations.

For species-habitat questions, the agentic configurations mainly relied on general ecological knowledge. When asked ``what is the structure of kingfisher’s habitat?'' they very generally discussed water networks, nesting banks, riparian vegetation, and possible anthropogenic barriers. The reactive configurations instead reported the output of the species-habitat module: approximately \(31.9\%\) of the basin was classified as suitable, and the habitat was characterised as moderately fragmented, with one dominant presence area and smaller satellite areas. The reactive response thus provided a reproducible spatial characterisation that was absent from the agentic answers.

Differences were also evident in scenario analysis. When asked ``assess ecosystem risk under a scenario with increased temperature and reduced tree cover'', the agentic configurations predicted a severe deterioration and provided detailed descriptions of possible hydroclimatic and ecological consequences, but did not derive their conclusions from explicit numerical projections because no projection was contained in the knowledge-base documents. The reactive agents, instead, reported the estimated change in high-risk cells, as returned by the scenario-projection module, and interpreted it. 

Likewise, when asked ``which intervention would be more effective between increasing tree cover and reducing land use?'', the agentic configurations introduced unsupported quantitative ranges for the expected effectiveness of the interventions. The reactive agents, instead, restricted the comparison to the model outputs and indicated that land-use reduction would have a stronger potential benefit on the ecosystem. 

Overall, these examples illustrate the principal architectural trade-off: agentic systems produced richer narratives, whereas reactive systems more consistently preserved the numerical and statistical evidence supplied by the DTE.

\subsection{Performance across evaluation dimensions}
\label{sec:performance_results}

Table~\ref{tab:performance} reports the assessments by the human and automated evaluators. Although the scores differed across individual dimensions, both evaluators identified broadly consistent strengths and weaknesses. In particular, they assigned low hallucination-control scores to the agentic configurations and substantially higher scores to all reactive configurations. The human evaluator assigned hallucination-control scores of \(33.64\%\) and \(40.23\%\) to agentic Llama~3.2 and Gemini, respectively, compared with values between \(77.50\%\) and \(89.30\%\) for their reactive counterparts. The automated evaluator reproduced this distinction, assigning \(30.00\%-32.00\%\) to the agentic configurations and \(82.50\%-89.30\%\) to the reactive ones. This was the clearest and most consistent architectural difference.

Both evaluators considered all configurations reasonably fluent. Fluency scores ranged from \(64.43\%\) to \(75.00\%\) for the human evaluator and from \(70.00\%\) to \(75.00\%\) for the automated evaluator. Thus, the stronger constraints imposed by the reactive architecture did not substantially compromise the linguistic quality of its answers.

Further similarities emerged in assessing correctness. Both evaluators assigned the highest correctness score to reactive Gemini (\(94.50\%\) and \(90.70\%\), respectively). More generally, the reactive configurations obtained higher correctness scores than the agentic configurations. Both evaluators also associated the agentic systems with a stronger scientific narrative or ecological storytelling. For example, the human expert assigned agentic Gemini an ecological-storytelling score of \(94.09\%\), whereas reactive-agent scores ranged from \(47.50\%\) to \(67.27\%\). The automated evaluator identified a less pronounced but similar tendency. Local ecological detail followed the same general pattern: the agentic configurations, particularly Gemini, incorporated more information specific to the Massaciuccoli ecosystem than the reactive systems.

Quantitative grounding exhibited the opposite pattern. The human evaluator assigned the agentic configurations only \(35.23\%-39.77\%\), compared with \(66.68\%-75.89\%\) for the reactive agents. By contrast, the automated evaluator was more favourable towards the agentic answers on this dimension, but still assigned the highest scores to the reactive configurations. The assessments supported the conclusion that performing the scientific calculations outside the LLM improved the connection between the generated answer and the underlying numerical evidence.

The overall agent scores were relatively close, ranging from \(61.20\%\) to \(68.59\%\) for the human evaluator and from \(64.14\%\) to \(68.04\%\) for the automated evaluator. All configurations provided answers of reasonable quality, but through different combinations of strengths. Agentic Gemini performed well overall, obtaining \(65.45\%\) from the human evaluator and \(66.71\%\) from the automated evaluator. Nevertheless, reactive Gemini achieved the highest overall score according to both evaluators, reaching \(68.59\%\) and \(68.04\%\), respectively. The reactive configuration retained Gemini's linguistic capabilities while improving correctness, quantitative grounding, and control over hallucinations.

The architectural effect was particularly evident for Llama~3.2~3B. Moving this model from the agentic to the reactive architecture increased its overall score from \(61.20\%\) to \(64.76\%\) according to the human evaluator and from \(64.14\%\) to \(67.30\%\) according to the automated evaluator. The improvement was especially pronounced for correctness, quantitative grounding, and hallucination control. Consequently, reactive Llama~3.2 reached performance comparable to configurations based on models with substantially greater computational capacity. 

Reactive Llama~4 Scout and Qwen3~8B occupied an intermediate position. Their overall scores remained above those of agentic Llama~3.2 but generally below those of the Gemini configuration. The human evaluator assigned them \(63.32\%\) and \(62.23\%\), respectively, while the automated evaluator assigned \(66.04\%\) and \(65.76\%\).

Overall, these findings suggest that model scale was not the primary determinant of performance within the reactive architecture. Once query interpretation, numerical processing, and model execution were delegated to specialised components, even a relatively small LLM such as Llama 3.2 could produce competitive answers. By contrast, the difference between the two agentic configurations was more pronounced: both evaluators rated agentic Gemini above agentic Llama~3.2. This observation suggests that model capability becomes more influential when the LLM must interpret retrieved information, infer relationships, and construct the answer without the query-specific analytical workflows.
%autonomously retrieve information, plan the analysis, interpret evidence, and construct the answer.

\subsection{Differences among overall scores}
\label{sec:configuration_differences}

The pairwise tests in Tables \ref{tab:performance} and \ref{tab:intraevaluatordifferences} indicate that the human evaluator identified systematic differences between every pair of configurations, despite the relatively narrow range of their overall average scores. 

The automated evaluator produced similar results for most comparisons but did not identify statistically significant overall differences between reactive Llama~4 Scout and Qwen3~8B. The automated scores were very close (\(66.04\%\) and \(65.76\%\), respectively), and so were the assessments of each dimension. Conversely, the human evaluator assessed that Scout performed better in ecological storytelling and hallucination control, whereas Qwen showed better use of local ecological detail.

The automated evaluator also assessed the score profiles of reactive Llama~3.2 (67.30\%) and reactive Gemini (68.04\%) as insufficiently different overall. Conversely, the human evaluator found a statistically significant difference between the configurations; their mean scores differed by 3.83 percentage points. According to the human evaluator, Gemini showed substantially higher correctness (\(94.50\%\) versus \(72.00\%\)), quantitative grounding (\(74.30\%\) versus \(66.68\%\)), hallucination control (\(89.30\%\) versus \(85.68\%\)), and fluency (\(74.80\%\) versus \(64.43\%\)). Instead, reactive Llama~3.2 obtained higher ecological storytelling (\(67.27\%\) versus \(56.80\%\)) and slightly higher local ecological detail (\(37.50\%\) versus \(31.80\%\)). 

Therefore, the automated evaluator did not capture fine details, particularly those related to correctness and quantitative grounding. This evaluator limited its assessment to the answer's resemblance to a correct, numerically grounded one and thus did not detect significant average differences across the evaluation dimensions. 

\subsection{Similarities and differences between evaluators}
\label{sec:evaluator_agreement}

The automated evaluator tended to assign slightly higher overall scores than the human expert (Table~\ref{tab:agreement}). The differences ranged from \(-0.5\) percentage points for reactive Gemini to \(3.5\) percentage points for reactive Qwen3, with an average difference of \(2.1\) percentage points across all configurations. Notably, the smallest discrepancy was observed for the best-performing configuration (reactive Gemini).

For each configuration, a two-sided Wilcoxon signed-rank test was applied across the answer scores assigned by the human and automated evaluators. None of these tests detected a statistically significant systematic difference between the two sets of scores: the configuration-specific \(p\)-values ranged from 0.16 to 0.46. A test conducted across all 42 paired configuration-dimension scores likewise did not detect a systematic difference (\(p=0.35\)). 

These results indicate that, despite the different perspectives on the individual dimensions, the automated evaluator did not exhibit a statistically detectable tendency to score the systems differently from the human expert. Together with the similar identification of the principal architectural strengths and weaknesses, this supports broad consistency between the two assessments.

The two evaluators differed on some individual dimensions, particularly the quantitative grounding of the answers. Generally, the automated evaluator did not detect distinctions that the human evaluator observed, owing to the human evaluator's detailed knowledge of the ecosystem and the underlying analytical evidence.  The results support the use of an automated evaluator for scalable comparative assessment rather than as a replacement for domain-expert review. Automated assessment was effective in reproducing the main performance patterns, whereas expert evaluation was fundamental for identifying finer ecological and interpretative differences.

\section{Discussion and conclusions}
\label{sec:discussion}

This study has presented a modular conversational digital twin of a natural ecosystem that integrates harmonised geospatial data, updated observations, scientific documents, species distribution modelling, ecosystem-risk assessment, and scenario simulation. In the proposed \textit{reactive} architecture, queries are routed to specialised modules that produce structured scientific evidence, while the LLM primarily extracts query parameters and communicates the results. The approach has been demonstrated for the Massaciuccoli Lake basin and compared with \textit{agentic}-AI baselines in which the LLM interpreted retrieved ecosystem information and previously generated model outputs to generate the answer. 

%The described DTE architectures offer interactive interfaces to improve mutual understanding among scientists, public authorities, citizens, and other stakeholders. They can adapt the presentation of evidence and recommendations to each user's knowledge, needs, and responsibilities. Synchronisation intervals may be longer than those of DTs representing physical objects, but the feedback remains bidirectional: observations update the virtual representation, while analyses inform regulations and conservation or planning interventions undertaken by authorities or citizens. These actions may modify the real ecosystem, and their effects would subsequently be incorporated into the DTE through new observations. However, because this loop may have substantial environmental and societal consequences, it is crucial that a DTE explicitly incorporates explainable reasoning.

The conversational interface enables scientists, public authorities, citizens, and other stakeholders to interrogate the ecosystem representation and explore conditional scenarios. In the proposed decision-support interpretation of a DTE, this interaction provides a pathway from digital analyses to potential actions in the physical ecosystem. Synchronisation intervals may be longer than those of DTs representing physical objects, but the feedback remains bidirectional and is human-mediated: authorities or other stakeholders may use the results to inform interventions and regulations, whose effects would subsequently be incorporated into the DTE through new observations. Therefore, because this loop may have substantial environmental and societal consequences, it is crucial that a DTE explicitly incorporates explainable reasoning. This pathway differs from automatic bidirectional control of an engineered asset.

The qualitative and quantitative comparisons revealed trade-offs between the architectures in this context. All configurations showed reasonably good fluency. Agentic configurations generated richer scientific narratives and ecological storytelling and made better use of local knowledge but frequently exceeded the available evidence, introducing hallucinated significance tests, scenario values, intervention effects, or causal interpretations, and showing weaker quantitative grounding. Reactive agents more consistently preserved the analytical results. They reported the conditions characterising high-risk areas, the direction and strength of correlations, SDM-derived habitat properties, and modelled scenario variations. Their answers were less elaborate but more closely connected to reproducible calculations. The evaluators assigned substantially higher scores to the reactive configurations for correctness, quantitative grounding, and hallucination control. 

Among the evaluated systems, reactive Gemini offered the best balance between these properties. Its agentic counterpart achieved lower yet reasonably good performance. In contrast, reactive Llama~3.2 achieved performance comparable to other reactive agents, including Scout and Qwen. Moreover, its substantially higher scores in correctness, quantitative grounding, and hallucination control than its agentic counterpart indicate that explicitly structured scientific reasoning can partially compensate for the LLM's limited capacity. Therefore, this architecture can potentially reduce the computational requirements of deploying conversational DTEs: when scientific reasoning is implemented externally, the LLM receives preprocessed evidence and performs a comparatively narrow linguistic task. Model capability becomes more important in the agentic architecture because the LLM must interpret the available evidence, develop an analytical narrative, and construct the response.

The overall performance assessments showed similar tendencies among evaluators, with some important differences. The automated evaluator did not identify significant overall differences between reactive Scout and Qwen or between reactive Llama~3.2 and Gemini, whereas these were evident to the human evaluator. The human assessment revealed finer trade-offs among correctness, storytelling, local detail, grounding, and hallucination control. Therefore, automated evaluation can support scalable comparison, but domain experts remain necessary for identifying subtle ecological and interpretative differences. 

It is worth noting that the similarity in overall scores should be interpreted as broadly consistent scoring tendencies within this benchmark, rather than evidence that the evaluators are interchangeable. A non-significant Wilcoxon test indicates that no systematic score difference was detected, but does not imply that the evaluators were equivalent. Overall averages can conceal dimension-specific discrepancies and different judgments about ecological interpretation or quantitative grounding. 

Generally, the overall scores, ranging from \(61.20\%\) to \(68.59\%\) for the human evaluator and from \(64.14\%\) to \(68.04\%\) for the automated evaluator, indicated reasonable rather than excellent performance. Improving the overall quality of the answers, therefore, remains a central objective for future development.

\subsection{Benefits of the reactive architecture}

The proposed framework complements the DTE approaches discussed in Section~\ref{sec:overview}. Data-oriented infrastructures provide access to harmonised and interoperable resources, but do not necessarily integrate them into an ecological interpretation. Model-oriented systems add simulations and scenario analysis but require substantial effort to connect heterogeneous models and develop user-oriented services. The evaluated LLM-led baseline offered flexible interpretation and narrative synthesis of retrieved evidence. The proposed reactive implementation additionally executed predefined scientific workflows and supplied their structured results to the language model. The observed performance differences support this integrated approach for the evaluated ecosystem questions, while comparisons with autonomous agents using equivalent analytical tools remain a separate research question. Generally, structuring results for the LLM improves explainability. Each answer can be traced to a query category, an analytical module, a set of input variables, and a structured result. Errors can consequently be located in routing, parameter extraction, model execution, contextual retrieval, or answer generation. 

The modules expose interpretable outputs, e.g., ranked stressors, correlation measures, predictor contributions, spatial habitat descriptors, explicit scenario modifications, and projected risk variations. Modularity allows the agent components to be supplemented or replaced independently with even more explainable models. Explainability could be enhanced through response curves, uncertainty intervals, local explanations, counterfactual analyses, provenance records, or more transparent statistical and ecological models without redesigning the complete agent. In this view, the reactive architecture should be preferred when answers communicate quantitative evidence relevant to environmental monitoring or management. Agentic reasoning remains useful for general conversation and broader ecological synthesis. The requirement for greater control also aligns with responsible AI principles such as transparency, traceability, accountability, reproducibility, and human oversight \citep{oecd2019ai,hleg2019trustworthyai}. The reactive architecture provides clearer points at which these principles can be implemented and audited.

Modularity also reduces implementation and operational costs. The currently supported query categories constitute a deliberately minimal set of ecologically relevant operations. However, since each operation is encapsulated in a specialised module, the framework can be extended by adding validated analytical components, routing examples, and answer templates without redesigning the overall architecture. Moreover, routing, prompting, answer generation, user interfaces, and containerised deployment can be reused across ecosystems. A new DTE can be produced essentially by configuring the study-area spatial extent, connecting its data and documents, adapting or retraining the ecosystem-specific models, and specialising the prompts, without modifying the remaining code. The performance of reactive Llama~3.2 indicates that smaller local LLMs may be sufficient, reducing hardware requirements and dependence on commercial services. Generalisation thus lowers the marginal cost of subsequent implementations, although data harmonisation, model validation, and expert supervision remain substantial activities.

\subsection{Implementation prerequisites}

Transferring the framework to another ecosystem requires the availability of

\begin{enumerate}
\item a harmonised geospatial dataset representing relevant environmental, climatic, biological, and anthropogenic conditions on compatible spatial and temporal supports;
\item updated observations from sensor networks, Earth-observation services, governmental repositories, biodiversity platforms, or authorised private sources;
\item a curated collection of scientific and technical documents describing the ecosystem, its services, pressures, biological communities, and management context;
\item a polygonal boundary and a common spatial grid for retrieving observations and aligning model outputs;
\item validated analytical models and baseline outputs, including an ecosystem-risk classification when risk-scenario modules are required;
\item sufficiently rich query examples, variable descriptions, metadata, and units to support routing and parameter extraction; and
\item storage, computing resources, an available LLM, and domain experts able to validate data, models, prompts, and answers.
\end{enumerate}

FAIRness and spatiotemporal harmonisation are fundamental because the agent must associate every result with identifiable, interoperable, and traceable evidence. 

Therefore, transferability here refers to architectural reusability: the modular framework can be adapted by replacing ecosystem-specific data, documentation, models, and configuration. However, comparable answer quality across ecosystems has not been empirically demonstrated. Differences in data availability, spatial scale, environmental dynamics, and modelling requirements may affect performance and require validation in each new application.

\subsection{Limitations and future work}

Results were based on a single ecosystem characterised by slowly changing conditions and comparatively rich data, supported by previously developed biodiversity and risk models. Portability was explained at the architectural level but was not empirically validated across ecosystems with different spatial scales, processes, and data availability. Moreover, the risk and scenario modules identified statistical associations rather than causal mechanisms because sufficiently long and homogeneous time series were unavailable. This was only a partial limitation because the case study demonstrated the agents' capabilities in the common situation in which spatially aggregated data are abundant but historical observations are limited. Moreover, the evaluation examined unsupported causal interpretations within correctness, quantitative grounding, and hallucination control. These assessments explicitly examined whether the generated answers respected the intended distinction between associations and causal effects.

Future work will apply the framework to other ecosystems, including the Baltic and Mediterranean Seas, for which extensive observations and ecosystem models are already available as FAIR data \citep{coro2023global,moll2024climate,pavirani2025assessing,Pavirani2026}. These applications will assess which components can be transferred directly and which require adaptation. The integration of longer time series, process-based models, and causal-inference methods will also be investigated to distinguish predictive associations from plausible cause-and-effect relationships.

A further limitation of the current study was that the 110-question benchmark covered the operations implemented in the DTE but did not capture the full variability of public interactions, extended conversations, ambiguous requests, or adversarial inputs. As the present study focused on describing and evaluating the software architecture, the assessment concentrated on answer quality as evidence of architectural effectiveness. Consequently, usability, accessibility, trust calibration, and effects on real management decisions were outside its scope. Future evaluations will include larger, more diverse query sets, multi-turn conversations, paraphrased and deliberately ambiguous questions, adversarial prompts, and requests requiring multiple modules. User studies involving scientists, monitoring authorities, policymakers, and citizens will evaluate whether the answers are understandable, appropriately trusted, and useful for practical environmental assessment and decision-making. This activity will also require changing the agent behaviour to better adapt the answers to the specific user group.

Another known limitation was that the evaluation involved one ecosystem expert and one LLM-based evaluator. Their broad consistency was encouraging, but did not capture the full range of disciplinary and stakeholder perspectives relevant to ecosystem management. Moreover, the human assessment provided expert judgment of the generated answers, rather than a benchmark of performance against expert-authored responses. Preparing detailed reference answers for the complete question set was beyond the expert's available effort. Future evaluation should include expert-authored reference answers and additional blinded assessments by experts without prior involvement in the project. Moreover, it should involve experts from complementary disciplines, including ecology, biodiversity, hydrology, climate science, environmental modelling, and territorial planning, as well as representatives of public authorities and local communities.

Finally, an additional limitation concerns the trade-off between flexibility and control in reactive agents. Answers may be comparatively rigid, and every additional analytical capability must be explicitly implemented, validated, and maintained. Moreover, constraining scientific computation does not guarantee high-quality answers. Although the LLM performs a narrower role than in the agentic architecture, the answer-building module must still harmonise heterogeneous module outputs with the user's intent, the retrieved ecological context, the conversational history, and the required scientific precautions. This complex integration may dilute or obscure otherwise correct analytical evidence and therefore represents a possible source of performance loss. 

The scientific modules implement established modelling approaches described and referenced in Section \ref{sec:modules}. Their methodological foundations support the analytical calculations, while the intermediate checks described in Section \ref{sec:evaluation_methodology} address their execution for the benchmark questions. A comprehensive component-level evaluation was thus outside the scope of this system-level comparison. Because the modules provide distinct scientific capabilities, removing one would primarily test the loss of that capability. Future controlled experiments will instead examine processing choices while retaining the required scientific functionality; for example, comparing automatic parameter extraction with manually verified parameters, or assessing documentary context while keeping analytical outputs fixed. Error tracing will further distinguish failures in extraction, analytical execution, and response generation. These developments should improve overall answer quality while further reducing, although not eliminating, hallucinations.

\subsection{Conclusions}
In conclusion, the Massaciuccoli case study demonstrated that natural-language interaction can provide practical access to heterogeneous ecosystem data and models. One main finding is that evidence-sensitive applications benefit from keeping scientific computation outside the LLM. This design improved correctness, quantitative grounding, traceability, and hallucination control while allowing a small LLM to achieve competitive performance. Agentic capabilities remain valuable for narrative synthesis, but they should operate within boundaries established by validated analytical components. The reactive architecture provides a reusable foundation for developing more reliable, explainable, and affordable conversational DTEs.

\section*{Acknowledgments}
This work has been conducted in the context of the OpenScience@Cnr FOE 2025 (ID DIT.AD022.242, CUP B53C25007660001) Project of the National Research Council of Italy. P.B. acknowledges financial support from the European Union’s Horizon Europe research and innovation programme through the GreenDIGIT project, under Grant Agreement No. 101131207. The authors wish to thank the Centro Servizi of CNR for providing computational machines to conduct the experiments, and the D4Science infrastructure \citep{assante2019enacting} for providing access to the Google Gemini Enterprise Agent Platform. Finally, the authors wish to thank Andrea Bertini, Andrea Fontanelli, and the Lega Italiana Protezione Uccelli (LIPU) for their involvement in system validation and for organising and participating in evaluation and exchange meetings over the past years.

\section*{Software and data availability}

The general software project framework for building conversational digital twins of ecosystems is openly available on GitHub at \\
\url{https://github.com/cybprojects65/EcosystemDigitalTwin}.

The code to build the agentic baselines is available on GitHub at \\
\url{https://github.com/cybprojects65/EcosystemDigitalTwinAgentic}

The Ollama-compatible bridge used to access Gemini, including its configuration and deployment instructions, is available within the EcosystemDigitalTwinAgentic GitHub repository under ``dte\_agentic/ollama-vertex-bridge''.

The information extraction prompts, which contain the related JSON schemas, are available in the following project subfolder, with the ``\_extraction'' suffix:\\
\url{https://github.com/cybprojects65/EcosystemDigitalTwin/tree/main/task_prompts}.

The reactive configurations in the EcosystemDigitalTwin repository can be changed through the \texttt{configuration/config.properties} file.

A separate, ready-to-execute, Dockerised Massaciuccoli demonstrator, containing the snapshot version of the agent used in the present study, is available at \url{https://github.com/cybprojects65/MassaciuccoliDigitalTwinReactive}.

Query examples for each task managed by the Massaciuccoli DTE are available at\\
\url{https://github.com/cybprojects65/EcosystemDigitalTwin/tree/main/query_examples}.

The report of all answers produced by the compared agents is available as a Microsoft Excel workbook at \\
\url{https://github.com/cybprojects65/EcosystemDigitalTwin/blob/main/benchmarking/Comparison_between_answers_by_all_Massaciuccoli_Agents.xlsx}.

Examples of queries and answers are available at \\
\url{https://github.com/cybprojects65/EcosystemDigitalTwin/blob/main/benchmarking/Agent_answers_examples.xlsx}

The full report of the assessments by the human and automatic evaluators is available at \\
\url{https://github.com/cybprojects65/EcosystemDigitalTwin/blob/main/benchmarking/Human_vs_automatic_query_level_report.xlsx}

The ChatGPT prompt used for the evaluation of the answers is available at \\
\url{https://github.com/cybprojects65/EcosystemDigitalTwin/blob/main/benchmarking/ChatGPT_prompt_for_evaluation.txt}

\printcredits

\bibliographystyle{cas-model2-names}

\bibliography{biblio}


%\newpage

\newpage
\section*{Figures and tables}
\clearpage

\clearpage

\clearpage

%\vspace{-0.5\baselineskip}
%\enlargethispage{2\baselineskip}

%\begin{table}[H]
 %   \centering
  %  \caption{Classification of species habitat coverage.}
   % \label{tab:habitat_coverage}
    %\begin{tabular}{ll}
     %   \hline
      %  Coverage \(C\) & Description \\
       % \hline
       % \(C < 5\%\)               & Very restricted habitat \\
       % \(5\% \leq C < 15\%\)     & Restricted habitat \\
       % \(15\% \leq C < 35\%\)    & Moderately widespread habitat \\
       % \(35\% \leq C < 60\%\)    & Widespread habitat \\
       % \(C \geq 60\%\)           & Very widespread habitat \\
       % \hline
   % \end{tabular}
%\end{table}

%\clearpage
\begin{table}[H]
    \centering
    \caption{Rule-based classification of habitat fragmentation, separation,
    connectivity, structure, and spread.}
    \label{tab:habitat_spatial_classification}
    \scriptsize
    \renewcommand{\arraystretch}{1.05}

    \begin{tabularx}{\textwidth}{
        p{2.5cm}
        p{4.8cm}
        X
    }
        \hline
        \textbf{Property} & \textbf{Condition} & \textbf{Classification} \\
        \hline

        Fragmentation
        & \(N_p \leq 1\)
        & Not fragmented \\

        & \(N_p > 1\), \(L \geq 60\%\), and \(D \leq 5\)
        & Low-to-moderate fragmentation with a dominant presence area \\

        & \(N_p > 1\), \(L \geq 40\%\), and not classified above
        & Moderate fragmentation with a dominant presence area and smaller
          presence areas \\

        & \(N_p > 1\), \(L < 40\%\), and
          (\(N_p > 20\) or \(M \leq 2\))
        & High fragmentation into many small presence areas \\

        & All other configurations
        & Moderate fragmentation \\
        \hline

        Core-to-satellite separation
        & \(D \leq 2\)
        & Nearby presence areas \\

        & \(2 < D \leq 5\)
        & Moderately separated presence areas \\

        & \(D > 5\)
        & Distant presence areas \\
        \hline

        Connectivity
        & \(N_p \leq 1\)
        & Continuous suitable habitat and presence area \\

        & \(N_p > 1\), \(L \geq 60\%\), and \(D \leq 5\)
        & Partly connected habitat dominated by one main presence area \\

        & \(N_p > 1\), \(L \geq 40\%\), and not classified above
        & Dominant presence area and smaller presence areas \\

        & \(N_p > 1\), \(L < 40\%\), and \(D > 5\)
        & Weakly connected or isolated presence areas \\

        & All other configurations
        & Moderately connected presence areas \\
        \hline

        Habitat structure
        & \(N_p \leq 1\)
        & Single core presence area \\

        & \(N_p > 1\) and \(L \geq 60\%\)
        & One dominant core presence area and smaller presence areas \\

        & \(N_p > 1\) and \(40\% \leq L < 60\%\)
        & Few main presence areas with additional smaller presence areas \\

        & \(N_p > 1\) and \(L < 40\%\)
        & Many distributed presence areas without a single dominant
          presence area \\
        \hline

        Habitat spread
        & \(C < 15\%\) and \(L \geq 60\%\)
        & Restricted and concentrated in one main presence area \\

        & \(C < 15\%\) and \(L < 60\%\)
        & Restricted but scattered \\

        & \(15\% \leq C < 35\%\) and \(L \geq 60\%\)
        & Moderately widespread but concentrated around one dominant
          presence area \\

        & \(15\% \leq C < 35\%\) and \(L < 60\%\)
        & Moderately widespread and distributed across several presence areas \\

        & \(C \geq 35\%\) and \(L \geq 60\%\)
        & Widespread but dominated by one main presence area \\

        & \(C \geq 35\%\) and \(L < 60\%\)
        & Widespread and distributed across multiple presence areas \\
        \hline
    \end{tabularx}

    \vspace{0.2cm}
    \begin{minipage}{0.98\textwidth}
        \footnotesize
        \(N_p\): number of connected presence areas;
        \(L\): percentage of the overall species presence contained in the largest presence area;
        \(D\): mean minimum border-to-border distance, expressed in cells;
        \(M\): median presence-area size, expressed in cells;
        \(C\): percentage of the study area classified as species presence.
        Core-to-satellite separation is defined only when \(N_p>1\).
        The rules within each property are evaluated sequentially.
    \end{minipage}

    \renewcommand{\arraystretch}{1}
\end{table}

\clearpage
\begin{table}[H]
    \centering
    \caption{Interpretation of the association measures used by the stressor correlations module. For Pearson's \(r\), the direction of the association is determined by the sign of the coefficient, whereas \(\eta\) and Cramér's \(V\) quantify association strength without indicating direction.}
    \label{tab:association_interpretation}
    \begin{tabular}{lll}
        \hline
        \textbf{Value range} &
        \textbf{Pearson correlation \(r\)} &
        \textbf{Correlation ratio \(\eta\) / Cramér's \(V\)} \\
        \hline
        \(0 \leq |r|,\eta,V < 0.10\) &
        Neutral &
        Negligible \\

        \(0.10 \leq |r|,\eta,V < 0.30\) &
        Positive or negative &
        Weak \\

        \(0.30 \leq |r|,\eta,V < 0.50\) &
        Moderately positive or negative &
        Moderate \\

        \(0.50 \leq |r|,\eta,V < 0.70\) &
        Strongly positive or negative &
        Strong \\

        \(0.70 \leq |r|,\eta,V \leq 1\) &
        Very strongly positive or negative &
        Very strong \\
        \hline
    \end{tabular}
\end{table}

\clearpage
\begin{table}[H]
    \centering
    \caption{Prompt-building strategies for the query categories supported by the proposed digital twin of the ecosystem.}
    \label{tab:prompt_strategies}
    \scriptsize
    \renewcommand{\arraystretch}{1.08}

    \begin{tabularx}{\textwidth}{
        >{\raggedright\arraybackslash}p{2.5cm}
        >{\raggedright\arraybackslash}p{5.0cm}
        >{\raggedright\arraybackslash}X
    }
        \hline
        \textbf{Category} &
        \textbf{Analytical evidence} &
        \textbf{Answer-building strategy} \\
        \hline

        Ecosystem drivers &
        Ranked features characterising high-risk cells and features with not distinctly characterising values &
        Explain the joint configuration associated with high-risk areas without interpreting the ranking as causal evidence or attributing risk to a single stressor. \\

        Stressor correlations &
        Numerical association values and qualitative strength and direction classes &
        Report the associations relevant to the requested source and target variables; distinguish correlation from causation and use the scientific context only to support ecological interpretation. \\

        Species habitat &
        Model diagnostics, predictor contributions, habitat coverage, fragmentation, connectivity, structure, and spread &
        Select the habitat properties requested by the user; report model reliability only when relevant and avoid presenting training AUC as independent ecological validation. \\

        Scenario projection &
        Description of the modified stressors and projected percentage variation in high-risk cells &
        Preserve the sign and magnitude of the projected variation, distinguish modified variables from model outcomes, and present the result as conditional on the simulated scenario. \\

        Scenario comparison &
        Descriptions and projected risk variations for scenarios \(A\) and \(B\) &
        Compare both projections against their common baseline and identify the relatively more or less adverse scenario without introducing variables or mechanisms absent from the evidence. \\

        Data retrieval &
        Names and processed values of the variables available in the live-data table &
        Select the variables relevant to the request, preserve their original names and units, and explicitly report when the requested information is unavailable. \\

        Conversation &
        Retrieved scientific context, when relevant, or the original query alone &
        Answer general questions, greetings, and clarification requests using the broader capabilities of the LLM, while requiring DTE-specific numerical claims to be supported by retrieved evidence. \\
        \hline
    \end{tabularx}

    \renewcommand{\arraystretch}{1}
\end{table}

\clearpage
\begin{table}[H]
  \centering
  \caption{Examples of answers generated by the agentic and reactive Massaciuccoli digital twins.}
  %\begin{adjustwidth}{-1cm}{-2cm}
    \begin{adjustbox}{max width=0.923\linewidth}
    \begin{tabular}{|l|p{8em}|r|l|p{68.96em}|p{22.71em}|}
    \multicolumn{1}{l}{\textbf{Query category}} & \multicolumn{1}{p{8em}}{\textbf{Query}} & \multicolumn{1}{l}{\textbf{Architecture}} & \multicolumn{1}{l}{\textbf{LLM}} & \multicolumn{1}{p{68.96em}}{\textbf{Answer}} & \multicolumn{1}{p{22.71em}}{\textbf{Notes}} \\
    \midrule
    \multicolumn{1}{|r|}{\multirow{6}[12]{*}{Ecosystem drivers}} & \multirow[t]{6}[12]{8em}{\raggedright How does biodiversity influence ecosystem risk?} & \multicolumn{1}{r|}{\multirow{2}[4]{*}{Agentic}} & Gemini 3.5 Flash & In the Massaciuccoli basin, biodiversity functions as the ecosystem's internal structural buffer and resilience engine. However, it does not influence risk in a simple, linear fashion. [..] Based on a multi-model analysis, biodiversity influences ecosystem risk through functional redundancy [..]; and lowering the risk [..]\newline{}Biodiversity metric changes are a primary indicator of escalating risk, particularly when specialized species begin to collapse. [..] & Rich ecological narrative, but largely generic and not demonstrably grounded in the module output. \\
\cmidrule{4-6}          &       &       & Llama 3.2 3B & Areas with high species richness tend to have moderate to high ecosystem risk. Species richness is positively correlated with ecosystem risk. We can perform hypothesis testing to determine whether the observed relationship between species richness and ecosystem risk is statistically significant. [..] The results of the hypothesis test indicate that the observed relationship between species richness and ecosystem risk is statistically significant (p $<$ 0.001) [..]\newline{}Recommendations: Implement conservation efforts to protect and restore habitats that support high species richness. Promote sustainable land use practices that minimize habitat destruction and fragmentation. Monitor and manage invasive species that can alter ecosystem processes and increase ecosystem risk. & Introduces a significance test not supported by the information shown and risks treating association as causation. \\
\cmidrule{3-6}          &       & \multicolumn{1}{r|}{\multirow{4}[8]{*}{Reactive}} & Gemini 3.5 Flash & Biodiversity is one of the key variables associated with high ecosystem risk in the Massaciuccoli basin, where it acts as a major factor within the main ranking of risk-related variables. High ecosystem risk is driven by the concurrency of high levels of several variables in specific areas, ordered by strength as land use increases, land imperviousness increases, grassland presence, grassland loss, tree cover loss, plant productivity, biodiversity, and temperature increases. The ecological context explains that ecosystem risk reflects interacting anthropogenic, climatic, biodiversity, vegetation, and land-use factors. Specifically, the principal risks emerge from the pressure of climate change (such as increasing aridity) on biodiversity-rich or productive areas, as well as from urban or industrial expansion. Climate change is projected to pose severe risks to biodiversity (such as bird species), alter the distribution of aquatic and terrestrial species as they search for cooler and water-richer areas, and significantly change the basin's biodiversity, which will likely have a negative impact on essential ecosystem services. & Strong grounding in the drivers-module output, although some resilience interpretation remains contextual. \\
\cmidrule{4-6}          &       &       & Llama 3.2 3B & The number of species potentially living in a cell is associated with high ecosystem risk. High species richness may indicate a more complex and resilient ecosystem, which can better withstand environmental stressors such as climate change, anthropogenic pressures, and hydrological dynamics. This association suggests that areas with high biodiversity may be more likely to exhibit intricate interactions among plant and animal communities, habitat structures, and ecological functions, potentially mitigating the impacts of human activities and climate-related drivers on the ecosystem. & Grounded in the detected association but adds a more speculative protective interpretation. \\
\cmidrule{4-6}          &       &       & Llama 4 Scout & The relationship between the number of species potentially living in the cell and ecosystem risk is poorly supported. This indicates that there is no direct discernible association between species richness and ecosystem risk in this context. Therefore, no ecological interpretation can be drawn about how species number influences ecosystem risk. & Very cautious, but may understate the high-risk feature evidence. \\
\cmidrule{4-6}          &       &       & Qwen3 8B & Biodiversity is associated with ecosystem risk through generic ecosystem interactions, but the available evidence does not support a clear or direct relationship. The reported dependency has low confidence and an unknown direction, indicating that no strong ecological interpretation can be drawn from the provided data. & Concise and cautious, but provides limited explanation of the ranked high-risk conditions. \\
    \midrule
    \multicolumn{1}{|r|}{\multirow{6}[12]{*}{Stressor correlations}} & \multirow[t]{6}[12]{8em}{\raggedright How does tree cover affect species richness?} & \multicolumn{1}{r|}{\multirow{2}[4]{*}{Agentic}} & Gemini 3.5 Flash & In the Massaciuccoli Lake basin and similar wetland ecosystems, tree cover is a fundamental driver of species richness. Its influence manifests through several critical ecological pathways: Habitat Complexity and Niche Availability Tree cover increases the structural complexity of the landscape, [..]; In the specific context of the Massaciuccoli wetland, tree cover plays a dual role in maintaining water quality: Filtering Effects: Vegetation acts as a natural buffer [..]; Species richness is closely tied to the ecosystem’s ability to resist and recover from disturbances [..]Summary for Ecosystem Risk Management: areas with low or decreasing tree cover density are consistently flagged as higher risk. This is because the loss of tree cover typically signifies a reduction in habitat quality, an increase in exposure to climatic stress, and a decline in the area's overall capacity to support diverse plant and animal populations. It only indicates that tree cover does not show a clear direct association with species richness in the reported correlative results. Its role may still emerge indirectly through interactions with other environmental variables and ecosystem processes. & Ecologically plausible, but contradicts the neutral direct relationship reported by the analytical evidence. \\
\cmidrule{4-6}          &       &       & Llama 3.2 3B & The relationship between tree cover and species richness is not directly represented as a factor driving changes in species richness. However, we can infer some potential relationships: Tree cover density: The average percentage of the area covered by trees (20.9) suggests that the tree cover density may be relatively high in certain areas. Grassland vegetation structure: Changes in grassland vegetation composition and structure may influence species richness in the surrounding areas, including the lake ecosystem. However, it's not clear how tree cover specifically affects species richness. It's possible that:\newline{}Tree cover provides habitat for certain species, potentially increasing species richness. Tree cover can alter local climate conditions, such as temperature and humidity, which may affect species distributions and abundance. Changes in tree cover density or composition may influence nutrient cycling, water quality, or other environmental factors that impact species richness. & More cautious than agentic Gemini, but relies mainly on generic hypotheses. \\
\cmidrule{3-6}          &       & \multicolumn{1}{r|}{\multirow{4}[8]{*}{Reactive}} & Gemini 3.5 Flash & Tree cover has an indirect relationship with species richness. In a complex ecosystem such as the one under analysis, species richness is unlikely to depend on tree cover alone. Instead, tree cover may interact with other ecological factors, such as habitat structure, vegetation productivity, microclimatic conditions, land use, grassland presence, and broader environmental pressures. Therefore, the neutral direct correlation should not be interpreted as evidence that tree cover is ecologically irrelevant. It only indicates that tree cover does not show a clear direct association with species richness in the reported correlative results. Its role may still emerge indirectly through interactions with other environmental variables and ecosystem processes. & Clearly separates the statistical result from broader ecological interpretation. \\
\cmidrule{4-6}          &       &       & Llama 3.2 3B & The density of tree cover appears to be somehow related to the number of species potentially living in the cell, with areas having high tree cover exhibiting lower numbers of species. This association may be attributed to the fact that dense vegetation can lead to increased competition for resources among plant and animal communities, potentially limiting biodiversity. Additionally, hydrological interactions between trees and aquatic organisms could also contribute to this relationship, as tree roots may alter water flow patterns or affect nutrient cycling, which in turn may impact species richness. & Uses the analytical direction but adds mechanisms that were not directly tested. \\
\cmidrule{4-6}          &       &       & Llama 4 Scout & The density of tree cover is negatively associated with the number of species potentially living in the cell, according to the model results. This relationship is characterised as weak and is classified as a vegetation-biodiversity interaction with moderate confidence.  & Concise and aligned with the structured correlation report. \\
\cmidrule{4-6}          &       &       & Qwen3 8B & Tree cover is associated with biodiversity, with a weak negative relationship. This suggests that areas with higher tree cover may have fewer species potentially present, though the strength of this association is low, and the confidence in this relationship is moderate. & Direct and appropriately cautious. \\
    \midrule
    \multicolumn{1}{|r|}{\multirow{6}[12]{*}{Species habitat}} & \multirow[t]{6}[12]{8em}{\raggedright What is the structure of kingfisher's habitat?} & \multicolumn{1}{r|}{\multirow{2}[4]{*}{Agentic}} & Gemini 3.5 Flash & The habitat connectivity for the Common Kingfisher (Alcedo atthis) in a wetland environment like the Massaciuccoli Lake basin is structured primarily by the spatial distribution of water resources, foraging perches, and suitable nesting banks. As an indicator species sensitive to environmental degradation, its movement and connectivity are defined by the following key ecological requirements: 1. Key Structural Drivers of Connectivity: Water Networks:  [..]; Foraging "Patches": [..]; Perch Availability: [..]; 2. Obstacles and Fragmentation: Connectivity in the Massaciuccoli basin is currently challenged by several factors: [..] Anthropogenic Barriers: [..]; Water Quality and Flow: [..]; Nesting Requirements: [..]; 3. Assessment and Conservation: In managed basins like Massaciuccoli, conservation efforts focus on maintaining these links by: Protecting Riparian Vegetation: [..];  Buffer Zones: [..]; Monitoring "Umbrella" Value: [..]. In short, for Alcedo atthis, connectivity is not just about physical space but about functional access to food and shelter.  & Detailed ecological narrative, but generic rather than based on the on-demand habitat model. \\
\cmidrule{4-6}          &       &       & Llama 3.2 3B & I'll provide some general insights based on available data and knowledge about Alcedo atthis (the Eurasian Kingfisher). Habitat connectivity:  Grassland and tree cover changes: grasslands have been relatively stable, with a 6.0\% gain in vegetation over the past decade (2025). Tree cover has also changed, with an average decrease of 3.0\% over the same period. Land use patterns: land is heavily exploited by human activities, with an average density of impervious surfaces of 2.3\%. Water availability: While not directly related to Alcedo atthis habitat, the data shows changes in water availability, including increased evaporation and decreased rainfall. Habitat connectivity structure: Fragmentation: The presence of impervious surfaces (2.3\%) and changes in land use patterns might suggest that habitats are fragmented, potentially isolating individual sites or groups. Isolation: The fact that grasslands have been relatively stable (6.0\% gain) could indicate that some areas may be isolated or disconnected from each other. Corridors and connectivity [..] would be necessary to maintain species movement and dispersal. Recommendations: Conducting field surveys [..]; Mapping [..]; Considering regional factors [...] & Uses local data, but the inferred habitat structure is not derived from an SDM spatial analysis. \\
\cmidrule{3-6}          &       & \multicolumn{1}{r|}{\multirow{4}[8]{*}{Reactive}} & Gemini 3.5 Flash & For the kingfisher (Alcedo atthis), suitable habitat is moderately widespread, covering about 31.9\% of the basin. The habitat pattern shows moderate fragmentation, with a dominant core presence area and smaller presence areas. The connectivity is characterised by a dominant presence area and smaller presence areas. The species is overall moderately widespread but concentrated around one dominant presence area. & Directly reports the SDM and spatial-analysis outputs without adding unsupported structure. \\
\cmidrule{4-6}          &       &       & Llama 3.2 3B & Habitat suitability for the kingfisher (Alcedo atthis) is mainly influenced by temperature and water presence, indicating sensitivity to key environmental gradients. Approximately 32\% of the territory represents a suitable core habitat. Suitable habitat appears widespread across the study area. Habitat connectivity is high, and suitable areas remain largely connected. Habitat structure is characterised by a dominant hotspot with secondary habitat nuclei. This connectivity allows individuals to move between habitats, maintaining genetic diversity and facilitating population dynamics. The presence of core habitats and hotspots suggests areas with high habitat quality and density that can support the species' ecological requirements. & Quantitatively grounded, although 'high connectivity' is stronger than the rule-based description shown for Gemini. \\
\cmidrule{4-6}          &       &       & Llama 4 Scout & Habitat suitability for Alcedo atthis is mainly influenced by temperature and water presence, indicating sensitivity to key environmental gradients. Approximately 32\% of the study area exceeds the hotspot threshold and represents highly suitable core habitat. Suitable habitat appears widespread across the study area, while highly suitable areas form an extensive core habitat. Habitat connectivity is high, and suitable areas remain largely connected. Habitat structure is characterised by a dominant hotspot with secondary habitat nuclei.  & Grounded in model outputs but uses a comparatively strong connectivity interpretation. \\
\cmidrule{4-6}          &       &       & Qwen3 8B & Habitat suitability for Alcedo atthis is mainly influenced by temperature and water presence, indicating sensitivity to key environmental gradients. Approximately 32\% of the study area exceeds the hotspot threshold and represents highly suitable core habitat. Suitable habitat appears widespread across the study area, while highly suitable areas form an extensive core habitat. Habitat connectivity is high, and suitable areas remain largely connected. Habitat structure is characterised by a dominant hotspot with secondary habitat nuclei. This suggests that the species' habitat is not highly fragmented, allowing for potential movement and interaction among individuals within the study area. & Clear and data-oriented, with little differentiation from the Scout response. \\
    \midrule
    \multicolumn{1}{|r|}{\multirow{6}[12]{*}{Scenario projection}} & \multirow[t]{6}[12]{8em}{\raggedright Assess ecosystem risk under a scenario with increased temperature and reduced tree cover.} & \multicolumn{1}{r|}{\multirow{2}[4]{*}{Agentic}} & Gemini 3.5 Flash & When a positive temperature anomaly couples with a reduction in tree cover, the Massaciuccoli basin passes a highly critical ecological inflexion point. This specific combination triggers a severe, non-linear compounding risk escalation. By simultaneously increasing the thermodynamic forcing (temperature) and dismantling the landscape’s primary protective armour (tree cover), the ecosystem is stripped of its buffering capacity. This pushes the spatial grid cells out of a manageable baseline directly into a volatile Critical Risk status. The compounding impact of these two stressors deconstructs the basin’s stability across four interconnected environmental vectors: The Hydro-Climatic Squeeze: Runaway PET and the Aridity Trap [..]; The Compound Effect: [..] The Vulnerability: [..] The wetland footprint rapidly contracts physically. Continuous aquatic networks fracture into isolated, shallow, overheated pools, accelerating habitat drying and leaving native wildlife with no climate refugia. [..] & Compelling narrative, but the severity and nonlinear tipping-point claims are unsupported by the displayed calculation. \\
\cmidrule{4-6}          &       &       & Llama 3.2 3B & Here is an assessment of ecosystem risk under a scenario with increased temperature and reduced tree cover: Increased Temperature: [..] +1.5°C; Potential impacts: Changes in species distributions and abundance due to altered thermal tolerance; Shifts in phenology [..], Increased risk of heat-related stress for plants and animals. Reduced Tree Cover: Percentage of the area covered by grassland: 15.3\%; Potential impacts: Reduced habitat for herbivores and other wildlife; Increased soil erosion due to reduced tree cover; Changes in local climate conditions, potentially affecting evapotranspiration rates. Combined Scenario: Ecosystem risk is likely to increase under this scenario due to the combined effects of increased temperature and reduced tree cover. The loss of tree cover will exacerbate the impacts of increased temperature on ecosystem functioning, leading to potential declines in biodiversity and ecosystem resilience. The changes in species distributions, abundance, and phenology may lead to disruptions in nutrient cycling, soil carbon sequestration, and other ecosystem processes that support biodiversity. [..] & Adds scenario values not supplied in the query and appears to confuse reduced tree cover with grassland information. \\
\cmidrule{3-6}          &       & \multicolumn{1}{r|}{\multirow{4}[8]{*}{Reactive}} & Gemini 3.5 Flash & The scenario with increased temperature and reduced tree cover would result in a 68.7\% overall ecosystem risk increase. Ecologically, medium-risk areas are characterised by rich biodiversity but a moderate reduction in tree cover and grassland. & Provides the requested quantitative projection, clearly grounded in the scenario module. \\
\cmidrule{4-6}          &       &       & Llama 3.2 3B & The scenario with increased temperature and reduced tree cover is expected to exacerbate ecosystem risk in the Massaciuccoli lake basin. The primary drivers of this increase, including species richness, tree cover density, vegetation productivity, change in grassland presence, and temperature change, are likely linked through ecological mechanisms such as altered nutrient cycling, changes in hydrological dynamics, and shifts in species distribution. For instance, reduced tree cover may lead to increased soil erosion and altered water quality, which can impact vegetation productivity and species richness.  Additionally, warmer temperatures may alter the distribution of aquatic organisms, leading to cascading consequences for ecosystem services and biodiversity. The secondary ecosystem responses suggest that these changes are not isolated events, but rather part of a complex web of interactions that require careful consideration in management planning. By understanding these ecological connections, policymakers can develop more effective strategies to mitigate ecosystem risk and promote resilience in the face of climate change. & Uses the scenario direction but devotes more space to inferred mechanisms than to the numerical result. \\
\cmidrule{4-6}          &       &       & Llama 4 Scout & The ecosystem risk assessment indicates risk increase under a scenario with increased temperature and reduced tree cover. The primary drivers are not specified, but secondary ecosystem responses show that species richness, tree cover density, and vegetation productivity are associated with the risk increase. These variables are likely linked to the ecosystem changes triggered by the increased temperature and reduced tree cover. & Quantitative and cautious about associations, but differs materially from the Gemini projection shown. \\
\cmidrule{4-6}          &       &       & Qwen3 8B & Under the scenario of increased temperature and reduced tree cover, the model identifies species richness as a primary driver of ecosystem risk, indicating a strong association with the predicted risk increase. Secondary responses include tree cover density, vegetation productivity, and changes in grassland presence, which are statistically linked to the scenario but do not imply directional changes in these variables. The association scores reflect variable importance within the model, not direct causality or measurable changes in environmental conditions. & Strong causal caution, but omits the main numerical scenario outcome. \\
    \midrule
    \multicolumn{1}{|r|}{\multirow{6}[12]{*}{Scenario comparison}} & \multirow[t]{6}[12]{8em}{\raggedright Which intervention would be more effective between increasing tree cover and reducing land use?} & \multicolumn{1}{r|}{\multirow{2}[4]{*}{Agentic}} & Gemini 3.5 Flash & Reducing land-use change is the more effective and urgent intervention. While increasing tree cover is a phenomenal localised buffering tool, it is ultimately a defensive strategy. Reducing land-use change targets the root cause of the ecosystem's instability. If you do not stop the ongoing conversion of natural marshes and peatlands into intensive agricultural matrices or urban sprawl, tree planting cannot scale quickly enough to compensate for the damage. Reducing land-use change essentially freezes the expansion of the ecosystem's primary threat vectors. It acts as an immediate containment strategy for the following systemic risks: [..]; The Role of Increasing Tree Cover (The Necessary Partner): [..] Increasing tree cover to a dense baseline (e.g., 70\%) is highly effective, but its true power is realised as a remediation tool within an already stabilised landscape. [..] The Limitation: [..] The Strength: [..] & Coherent management narrative, but causal priority is asserted rather than calculated from the DTE. \\
\cmidrule{4-6}          &       &       & Llama 3.2 3B & Reducing land-use change might be a more effective intervention than increasing tree cover in the long run. Land-use change is a major driver of ecosystem degradation: changes in land use, such as deforestation, urbanisation, and agricultural expansion, are often the primary cause. By reducing these changes, you can prevent or slow down the degradation process. Tree cover can be restored, but land-use changes cannot be reversed: Once land is converted to a different land use (e.g., from forest to agriculture), it's difficult to revert back to its original state. Reducing land-use change can help maintain ecosystem integrity and reduce the need for costly restoration efforts. Reducing land-use change has broader benefits: you can also reduce greenhouse gas emissions, protect water resources, and promote sustainable development. Increasing tree cover alone might lead to a 10-30\% reduction in ecosystem risk. Reducing land-use change alone might increase ecosystem risk reduction by 20-50\% or more. This is because reducing land-use change can prevent or slow down ecosystem degradation, while increasing tree cover may only mitigate some of the effects. Combined effects: Tree cover + reduced land-use change: Combining both strategies could lead to increased ecosystem risk reduction [..] & Reasonable qualitative conclusion, but includes invented quantitative estimates. \\
\cmidrule{3-6}          &       & \multicolumn{1}{r|}{\multirow{4}[8]{*}{Reactive}} & Gemini 3.5 Flash & Land-use change has a stronger influence on ecosystem risk than tree cover. Based on the model outputs, land-use change is more strongly associated with ecosystem risk than tree cover density. This association is supported by evidence indicating that changes in land use and vegetation cover, particularly through anthropisation and the expansion of artificially shaped territories, act as primary harmful forces that drive ecosystem risk and affect biodiversity. Conversely, tree cover density exhibits a lower model sensitivity within the current prediction, indicating a weaker relative association with the evaluated ecosystem risk. & Grounded in comparative model output and avoids invented effect sizes. \\
\cmidrule{4-6}          &       &       & Llama 3.2 3B & Land-use change has a stronger influence on ecosystem risk than tree cover. This suggests that changes in land use, such as deforestation or urbanisation, are having a more significant impact on the overall health and resilience of ecosystems, potentially leading to increased biodiversity loss, reduced ecosystem services, and altered hydrological cycles. & Direct and evidence-based, with a limited generic discussion of wider ecosystem effects. \\
\cmidrule{4-6}          &       &       & Llama 4 Scout & In the current model, land-use change shows a stronger influence on ecosystem risk than tree cover. This suggests that changes in land use, such as urban expansion or agricultural development, have a more significant impact on ecosystem risk than the density of tree cover. Therefore, interventions targeting land-use change, such as sustainable land-use planning or conservation efforts, may be more effective in mitigating ecosystem risk. & Concise and appropriately conditional on the model. \\
\cmidrule{4-6}          &       &       & Qwen3 8B & In the current model, land-use change shows a stronger influence on ecosystem risk than tree cover, indicating that changes in land use are more closely linked to the observed risk levels in the Massaciuccoli lake basin. Tree cover, while relevant, shows a much weaker influence on the modelled ecosystem risk, suggesting that its role in driving the assessed risk is less prominent within this specific context. & Clear and restrained interpretation of the comparative evidence. \\
    \bottomrule
    \end{tabular}%
    \end{adjustbox}
   % \end{adjustwidth}
  \label{tab:examples}%
\end{table}%

\clearpage
\begin{table}[H]
  \centering
  \caption{Performance comparison between two agentic-AI configurations and four reactive-AI configurations for the Massaciuccoli digital twin of the ecosystem. The performance was evaluated across 7 dimensions by two evaluators: a \textit{human} expert familiar with the ecosystem and an \textit{automated} expert simulated with ChatGPT 5.5. Colours represent overall scores on a red-to-green gradient, from lowest to highest. Percentages are given to two decimal places to better highlight fine differences between the evaluations.}
  \begin{adjustbox}{max width=\textwidth}
    \begin{tabular}{rp{9.835em}rrrrrr}
\cmidrule{3-8}          & \multicolumn{1}{r|}{} & \multicolumn{2}{c|}{\textbf{Agentic}} & \multicolumn{4}{c|}{\textbf{Reactive}} \\
    \midrule
    \multicolumn{1}{|l|}{\textbf{Evaluator}} & \multicolumn{1}{l|}{\textbf{Evaluation dimension}} & \multicolumn{1}{l|}{\textbf{Llama 3.2 3B}} & \multicolumn{1}{l|}{\textbf{Gemini 3.5 Flash}} & \multicolumn{1}{l|}{\textbf{Llama 4 Scout}} & \multicolumn{1}{l|}{\textbf{Qwen3 8B}} & \multicolumn{1}{l|}{\textbf{Llama 3.2 3B}} & \multicolumn{1}{l|}{\textbf{Gemini 3.5 Flash}} \\
    \midrule
    \midrule
    \multicolumn{1}{c}{\multirow{7}[2]{*}{Human}} & Correctness & 57.95\% & 63.64\% & 70.55\% & 72.73\% & 72.00\% & 94.50\% \\
          & Scientific narrative & 70.45\% & 75.00\% & 59.09\% & 54.55\% & 59.77\% & 58.60\% \\
          & Quantitative grounding & 39.77\% & 35.23\% & 74.66\% & 75.89\% & 66.68\% & 74.30\% \\
          & Ecological storytelling & 75.00\% & 94.09\% & 52.95\% & 47.50\% & 67.27\% & 56.80\% \\
          & Hallucination control & 33.64\% & 40.23\% & 82.05\% & 77.50\% & 85.68\% & 89.30\% \\
          & Local ecological detail & 76.59\% & 75.00\% & 37.50\% & 41.36\% & 37.50\% & 31.80\% \\
          & Fluency & 75.00\% & 75.00\% & 66.48\% & 66.07\% & 64.43\% & 74.80\% \\
    \midrule
    \multicolumn{1}{c}{\multirow{7}[2]{*}{Automated}} & \multicolumn{1}{l}{Correctness} & 72.00\% & 70.00\% & 83.40\% & 81.60\% & 90.20\% & 90.70\% \\
          & \multicolumn{1}{l}{Scientific narrative} & 78.00\% & 74.00\% & 59.50\% & 63.90\% & 59.30\% & 58.60\% \\
          & \multicolumn{1}{l}{Quantitative grounding} & 66.00\% & 71.00\% & 78.00\% & 73.90\% & 75.20\% & 74.30\% \\
          & \multicolumn{1}{l}{Ecological storytelling} & 76.00\% & 72.00\% & 54.10\% & 53.60\% & 62.00\% & 56.80\% \\
          & \multicolumn{1}{l}{Hallucination control} & 32.00\% & 30.00\% & 84.10\% & 84.10\% & 82.50\% & 89.30\% \\
          & \multicolumn{1}{l}{Local ecological detail} & 55.00\% & 75.00\% & 29.10\% & 28.40\% & 30.50\% & 31.80\% \\
          & \multicolumn{1}{l}{Fluency} & 70.00\% & 75.00\% & 74.10\% & 74.80\% & 71.40\% & 74.80\% \\
    \midrule
          & \textbf{Expert avg.} & \cellcolor[rgb]{ .973,  .412,  .42}61.20\% & \cellcolor[rgb]{ .812,  .871,  .51}65.45\% & \cellcolor[rgb]{ .992,  .792,  .49}63.32\% & \cellcolor[rgb]{ .98,  .592,  .455}62.23\% & \cellcolor[rgb]{ .906,  .894,  .514}64.76\% & \cellcolor[rgb]{ .388,  .745,  .482}68.59\% \\
          & \textbf{Automated avg.} & \cellcolor[rgb]{ .973,  .412,  .42}64.14\% & \cellcolor[rgb]{ .878,  .886,  .514}66.71\% & \cellcolor[rgb]{ .992,  .843,  .502}66.04\% & \cellcolor[rgb]{ .992,  .776,  .49}65.76\% & \cellcolor[rgb]{ .663,  .827,  .502}67.30\% & \cellcolor[rgb]{ .388,  .745,  .482}68.04\% \\
    \end{tabular}%
    \end{adjustbox}
  \label{tab:performance}%
\end{table}%

\clearpage
\begin{table}[H]
  \centering
  \caption{Results of the pairwise two-sided Wilcoxon signed-rank tests at a significance level of 0.05 to highlight significantly different average scores across the compared agent configurations and architectures. The pairwise assessments are reported separately for each evaluator. Green indicates significantly different scores (p-value $<$ 0.05). Red indicates non-significantly different scores (p-value $>$ 0.05).}
  \begin{adjustbox}{max width=\textwidth}
    \begin{tabular}{|c|l|r|r|r|r|r|r|}
    \toprule
    \multicolumn{8}{|c|}{Human expert - Pairwise differences in scores} \\
    \midrule
    \midrule
    \multicolumn{1}{r}{} &       & \multicolumn{2}{c|}{Agentic} & \multicolumn{4}{c|}{Reactive} \\
\cmidrule{3-8}    \multicolumn{1}{r}{} &       & \multicolumn{1}{l|}{Llama 3.2 3B} & \multicolumn{1}{l|}{Gemini 3.5 Flash} & \multicolumn{1}{l|}{Llama 4 Scout} & \multicolumn{1}{l|}{Qwen3 8B} & \multicolumn{1}{l|}{Llama 3.2 3B} & \multicolumn{1}{l|}{Gemini 3.5 Flash} \\
    \midrule
    \multirow{2}[4]{*}{Agentic} & Llama 3.2 3B & \cellcolor[rgb]{ .816,  .816,  .816} & \cellcolor[rgb]{ .855,  .949,  .816} & \cellcolor[rgb]{ .855,  .949,  .816} & \cellcolor[rgb]{ .855,  .949,  .816} & \cellcolor[rgb]{ .855,  .949,  .816} & \cellcolor[rgb]{ .855,  .949,  .816} \\
\cmidrule{2-8}          & Gemini 3.5 Flash &       & \cellcolor[rgb]{ .816,  .816,  .816} & \cellcolor[rgb]{ .855,  .949,  .816} & \cellcolor[rgb]{ .855,  .949,  .816} & \cellcolor[rgb]{ .855,  .949,  .816} & \cellcolor[rgb]{ .855,  .949,  .816} \\
    \midrule
    \multirow{4}[8]{*}{Reactive} & Llama 4 Scout &       &       & \cellcolor[rgb]{ .816,  .816,  .816} & \cellcolor[rgb]{ .855,  .949,  .816} & \cellcolor[rgb]{ .855,  .949,  .816} & \cellcolor[rgb]{ .855,  .949,  .816} \\
\cmidrule{2-8}          & Qwen3 8B &       &       &       & \cellcolor[rgb]{ .816,  .816,  .816} & \cellcolor[rgb]{ .855,  .949,  .816} & \cellcolor[rgb]{ .855,  .949,  .816} \\
\cmidrule{2-8}          & Llama 3.2 3B &       &       &       &       & \cellcolor[rgb]{ .816,  .816,  .816} & \cellcolor[rgb]{ .855,  .949,  .816} \\
\cmidrule{2-8}          & Gemini 3.5 Flash &       &       &       &       &       & \cellcolor[rgb]{ .816,  .816,  .816} \\
    \midrule
    \multicolumn{8}{|c|}{Automated expert - Pairwise differences in scores} \\
    \midrule
    \midrule
    \multicolumn{1}{r}{} &       & \multicolumn{2}{c|}{Agentic} & \multicolumn{4}{c|}{Reactive} \\
\cmidrule{3-8}    \multicolumn{1}{r}{} &       & \multicolumn{1}{l|}{Llama 3.2 3B} & \multicolumn{1}{l|}{Gemini 3.5 Flash} & \multicolumn{1}{l|}{Llama 4 Scout} & \multicolumn{1}{l|}{Qwen3 8B} & \multicolumn{1}{l|}{Llama 3.2 3B} & \multicolumn{1}{l|}{Gemini 3.5 Flash} \\
    \midrule
    \multirow{2}[4]{*}{Agentic} & Llama 3.2 3B & \cellcolor[rgb]{ .816,  .816,  .816} & \cellcolor[rgb]{ .855,  .949,  .816} & \cellcolor[rgb]{ .855,  .949,  .816} & \cellcolor[rgb]{ .855,  .949,  .816} & \cellcolor[rgb]{ .855,  .949,  .816} & \cellcolor[rgb]{ .855,  .949,  .816} \\
\cmidrule{2-8}          & Gemini 3.5 Flash &       & \cellcolor[rgb]{ .816,  .816,  .816} & \cellcolor[rgb]{ .855,  .949,  .816} & \cellcolor[rgb]{ .855,  .949,  .816} & \cellcolor[rgb]{ .855,  .949,  .816} & \cellcolor[rgb]{ .855,  .949,  .816} \\
    \midrule
    \multirow{4}[8]{*}{Reactive} & Llama 4 Scout &       &       & \cellcolor[rgb]{ .816,  .816,  .816} & \cellcolor[rgb]{ 1,  0,  0} & \cellcolor[rgb]{ .855,  .949,  .816} & \cellcolor[rgb]{ .855,  .949,  .816} \\
\cmidrule{2-8}          & Qwen3 8B &       &       &       & \cellcolor[rgb]{ .816,  .816,  .816} & \cellcolor[rgb]{ .855,  .949,  .816} & \cellcolor[rgb]{ .855,  .949,  .816} \\
\cmidrule{2-8}          & Llama 3.2 3B &       &       &       &       & \cellcolor[rgb]{ .816,  .816,  .816} & \cellcolor[rgb]{ 1,  0,  0} \\
\cmidrule{2-8}          & Gemini 3.5 Flash &       &       &       &       &       & \cellcolor[rgb]{ .816,  .816,  .816} \\
    \bottomrule
    \end{tabular}%
    \end{adjustbox}
  \label{tab:intraevaluatordifferences}%
\end{table}%

\clearpage
\begin{table}[H]
  \centering
  \caption{Differences between the mean scores assigned by the automated and human evaluators (in percentage points) across the compared agent configurations and architectures. The Wilcoxon p-values are reported to quantify the degree of discrepancy, with light-red colour highlighting that the discrepancies are non-significant. The average overall difference between the evaluators was calculated as the difference between the average scores for each evaluator across the assessments.}
  \label{tab:agreement}%
  \begin{adjustbox}{max width=\textwidth}
    \begin{tabular}{|r|l|r|r|}
\cmidrule{3-4}    \multicolumn{1}{r}{} &       & \multicolumn{1}{p{15.415em}|}{Automated score - Expert score (\%pt)} & \multicolumn{1}{p{7.5em}|}{Two-sided Wilcoxon p-values ($\alpha=0.05$)} \\
    \midrule
    \multicolumn{1}{|c|}{\multirow{2}[4]{*}{Agentic}} & Llama 3.2 3B & 2.9   & \cellcolor[rgb]{ .984,  .886,  .835} 0.20 \\
\cmidrule{2-4}          & Gemini 3.5 Flash & 1.3   & \cellcolor[rgb]{ .984,  .886,  .835} 0.45 \\
    \midrule
    \multicolumn{1}{|c|}{\multirow{4}[8]{*}{Reactive}} & Llama 4 Scout & 2.7   & \cellcolor[rgb]{ .984,  .886,  .835} 0.24 \\
\cmidrule{2-4}          & Qwen3 8B & 3.5   & \cellcolor[rgb]{ .984,  .886,  .835} 0.16 \\
\cmidrule{2-4}          & Llama 3.2 3B & 2.5   & \cellcolor[rgb]{ .984,  .886,  .835} 0.27 \\
\cmidrule{2-4}          & Gemini 3.5 Flash & -0.5  & \cellcolor[rgb]{ .984,  .886,  .835} 0.46 \\
    \midrule
          & Average overall difference & 2.1   & \cellcolor[rgb]{ .984,  .886,  .835} 0.35 \\
    \bottomrule
    \end{tabular}%
    \end{adjustbox}
\end{table}%

\end{document}